\documentclass{article}

\usepackage{hankel}

\begin{document}

\title{Exploiting Hankel--Toeplitz Structures \\ for Fast Computation of Kernel Precision Matrices}




\author{\name Frida Viset~$\dagger$\email f.m.viset@tudelft.nl \\
      \addr Delft Center for Systems and Control\\
      Delft University of Technology, The Netherlands
      \AND
      \name Anton Kullberg~$\dagger$ \email anton.kullberg@liu.se \\
      \addr Department of Electrical Engineering\\
      Link\"oping University, Sweden
      \AND
      \name Frederiek Wesel \email f.wesel@tudelft.nl\\
      \addr Delft Center for Systems and Control\\
      Delft University of Technology, The Netherlands
      \AND
      \name Arno Solin \email arno.solin@aalto.fi\\
      \addr Department of Computer Science\\
      Aalto University, Finland}


\newcommand{\fix}{\marginpar{FIX}}
\newcommand{\new}{\marginpar{NEW}}

\def\month{07}  
\def\year{2024} 
\def\openreview{\url{https://openreview.net/forum?id=s9ylaDLvdO}} 



\newcommand\note[1]{\textcolor{red}{#1}}

\renewcommand{\cite}[1]{\textcolor{red}{DO NOT USE CITE (use citep/citet instead)}}
\renewcommand{\ref}[1]{\textcolor{red}{AVOID REF (prefer CREF)}}

\newcommand{\arno}[1]{\textcolor{blue}{\bf [#1]}}

\newcommand\blfootnote[1]{%
  \begingroup
  \renewcommand\thefootnote{}\footnote{#1}%
  \addtocounter{footnote}{-1}%
  \endgroup
}

\maketitle

\begin{abstract}
    The \gls{hgp} approach offers a hyperparameter-independent basis function approximation for speeding up \gls{gp} inference by projecting the \gls{gp} onto $M$ basis functions. These properties result in a favorable data-independent $\mathcal{O}(M^3)$ computational complexity during hyperparameter optimization but require a dominating one-time precomputation of the precision matrix costing $\mathcal{O}(NM^2)$ operations. In this paper, we lower this dominating computational complexity to $\mathcal{O}(NM)$ with \emph{no additional approximations}. We can do this because we realize that the precision matrix can be split into a sum of Hankel--Toeplitz matrices, each having $\mathcal{O}(M)$ unique entries. Based on this realization we propose computing only these unique entries at $\mathcal{O}(NM)$ costs. 
    Further, we develop two theorems that prescribe sufficient conditions for the complexity reduction to hold generally for a wide range of other approximate \gls{gp} models, such as the \gls{vff} approach.
    The two theorems do this with no assumptions on the data and no additional approximations of the \gls{gp} models themselves.
    Thus, our contribution provides a pure speed-up of several existing, widely used, \gls{gp} approximations, \emph{without further approximations}.\looseness-1
\end{abstract}

\blfootnote{$\dagger$-These authors contributed equally to this work. A reference implementation built on top of GPJax is available at \href{https://github.com/AOKullberg/hgp-hankel-structure}{https://github.com/AOKullberg/hgp-hankel-structure}.}

\section{Introduction}\label{sec:introduction}

\emph{Gaussian Processes} \citep[\gls{gp}s,][]{rasmussen2005gaussian} provide a flexible formalism for modeling functions which naturally allows for the incorporation of prior knowledge and the production of uncertainty estimates in the form of a predictive distribution. 
Typically a \gls{gp} is instantiated by specifying a prior mean and covariance (kernel) function, which allows for incorporation of prior knowledge. When data becomes available, the \gls{gp} can then be conditioned on the observations, yielding a new \gls{gp} which can be used for predictions and uncertainty quantification.

While all these operations have closed-form expressions for regression, the computations of the mean and covariance of the predictive \gls{gp} require instantiating and inverting the kernel (Gram) matrix, which encodes pair-wise similarities between all data. 
These operations require respectively $\mathcal{O}(N^2)$ and $\mathcal{O}(N^3)$ computations, where $N$ is the number of data points.
Furthermore, if one wishes to optimize the hyperparameters of the kernel function, which in \glspl{gp} is typically accomplished by optimizing the log-marginal likelihood, the kernel matrix needs to be instantiated and inverted multiple times, further hindering the applicability of \glspl{gp} to large-scale datasets.

\begin{figure}[t]
    \centering\scriptsize    
    \begin{tikzpicture}
    \newcommand\spread[2][]{%
    \addplot [thick, mark=none, color=#1]
        table [%
        col sep=comma,
        x={m}, y={t_median}]
        {\currfiledir /../data/#2.csv};
    \addplot [thick, name path=upper, mark=none, color=#1, dashed, forget plot, opacity=.75]
        table [%
        col sep=comma,
        x={m}, y={t_max}] %
        {\currfiledir /../data/#2.csv};
    \addplot [thick, name path=lower, mark=none, color=#1, dashed, forget plot, opacity=.75]
        table [%
        col sep=comma,
        x={m}, y={t_min}] %
        {\currfiledir /../data/#2.csv};
    \addplot[color=#1, fill opacity=0.15, forget plot] fill between[of=lower and upper];
    }
        \begin{semilogyaxis}[
        axis lines=left,
        axis background style={fill=black!02,rounded corners=2pt},
        height=.33\textwidth, 
        width=.5\textwidth,
        grid=major,
        ytick={0.01,0.1,1,10,100,1000,10000},
        yticklabels={{},{$10^{-1}$},{},{$10^1$},{},{$10^3$},{}},
        ymin=0.1,
        legend entries={\textsc{hgp}, Ours, $\mathcal{O}(M^2)$, $\mathcal{O}(M)$},
        legend columns=4,
        legend style={
        at={(0.5, 0.0)},
        anchor=south,
        draw=none,
        fill=black!02
        },
        ylabel=$\leftarrow$~Wall-clock time (s),
        scaled x ticks=base 10:-3,
        xlabel={Number of \textsc{bf}s, $M~({\times}10^3)$},
        xtick scale label code/.code={},
        y tick label style={font=\tiny},
        x tick label style={font=\tiny},
        ]
        \spread[Paired-B]{HGPtiming}
        \spread[Paired-H]{SHGPtiming}
        \addplot [thick, samples=100, color=gray, domain=0:59000, dashed, opacity=1] {3 + 8e-7 * x^2};
        \addplot [thick, samples=100, color=gray, domain=0:59000, dotted, opacity=1] {1e-2 + 2.5e-4 * x};
        \end{semilogyaxis}
    \end{tikzpicture}\\[-6pt]
    \caption{\textbf{An order of magnitude speed-up without any additional approximations:} Wall-clock time to compute the precision matrix for an increasing number $M$ of basis functions.} 
    \label{fig:timing}
    \vspace*{-6pt}
\end{figure}
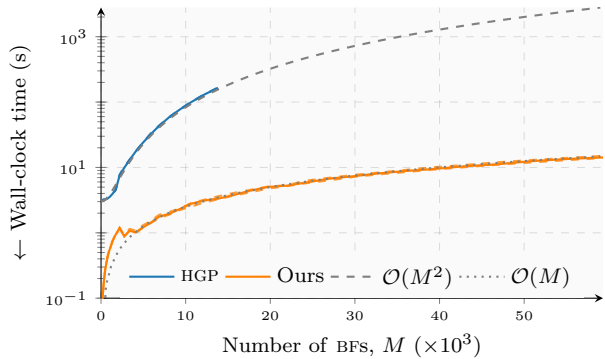

The ubiquitous strategy to reduce this computational complexity consists in approximating the kernel matrix in terms of \glspl{bf}, yielding what is essentially a low-rank or ``sparse'' approximation of the kernel function \citep[see][]{rasmussen2005gaussian,quinonero-candelaUnifyingViewSparse2005,snelsonSparseGaussianProcesses2005}.
Computational savings can then be achieved by means of the matrix inversion lemma, which requires instantiating and inverting the \emph{precision matrix} (sum of the outer products of the \glspl{bf}) instead of the kernel matrix at prediction, thereby lowering the computational complexity of inference to $\mathcal{O}(NM^2 + M^3)$, where $M$ is the number of \glspl{bf}.
If the number of \glspl{bf} is chosen smaller than the number of data points in the training set ($M<N$), computational benefits arise and the computational costs for hyperparameter optimization and inference are dominated by $\mathcal{O}(NM^2)$.

It is less widely known that this cost can be improved further.
A notable \gls{bf} framework is the \gls{hgp} \citep{solinHilbertSpaceMethods2020} which projects the \gls{gp} onto a dense set of orthogonal, hyperparameter-independent \glspl{bf}. 
This deterministic approximation is particularly attractive as it exhibits fast convergence guarantees to the full \gls{gp} in terms of the number of \glspl{bf} for smooth shift-invariant kernels compared to other approximations. 
Furthermore, the fact that the \glspl{bf} are hyperparameter-independent speeds up \gls{gp} hyperparameter optimization considerably, which in the \gls{hgp} requires only $\mathcal{O}(M^3)$ operations after a one-time precomputation of the precision matrix, costing $\mathcal{O}(NM^2)$. 
These favorable properties have ensured a relatively wide usage of the \gls{hgp}~\citep[see e.g.][]{svenssonComputationallyEfficientBayesian2016,berntorpOnlineBayesianInference2021a,kokMagneticFieldSLAM2018}, and it is available in, e.g.,
PyMC \citep{pymc} and Stan \citep{riutortPractical2022}. 
However, as argued by \citet{lindgrenSpde2022}, a high number of \glspl{bf} may be required for a faithful approximation of the full model. Thus, the \gls{hgp} is typically  employed in applications where forming the initial $\mathcal{O}(NM^2)$ projection does not become too heavy.\looseness-1

\begin{figure}[t]
    \captionsetup[subfigure]{labelformat=empty}
       \centering
      \begin{subfigure}[1D]{0.16\textwidth}
           \centering
          \stackunder[5pt]{1D}{\includegraphics[width=\textwidth]{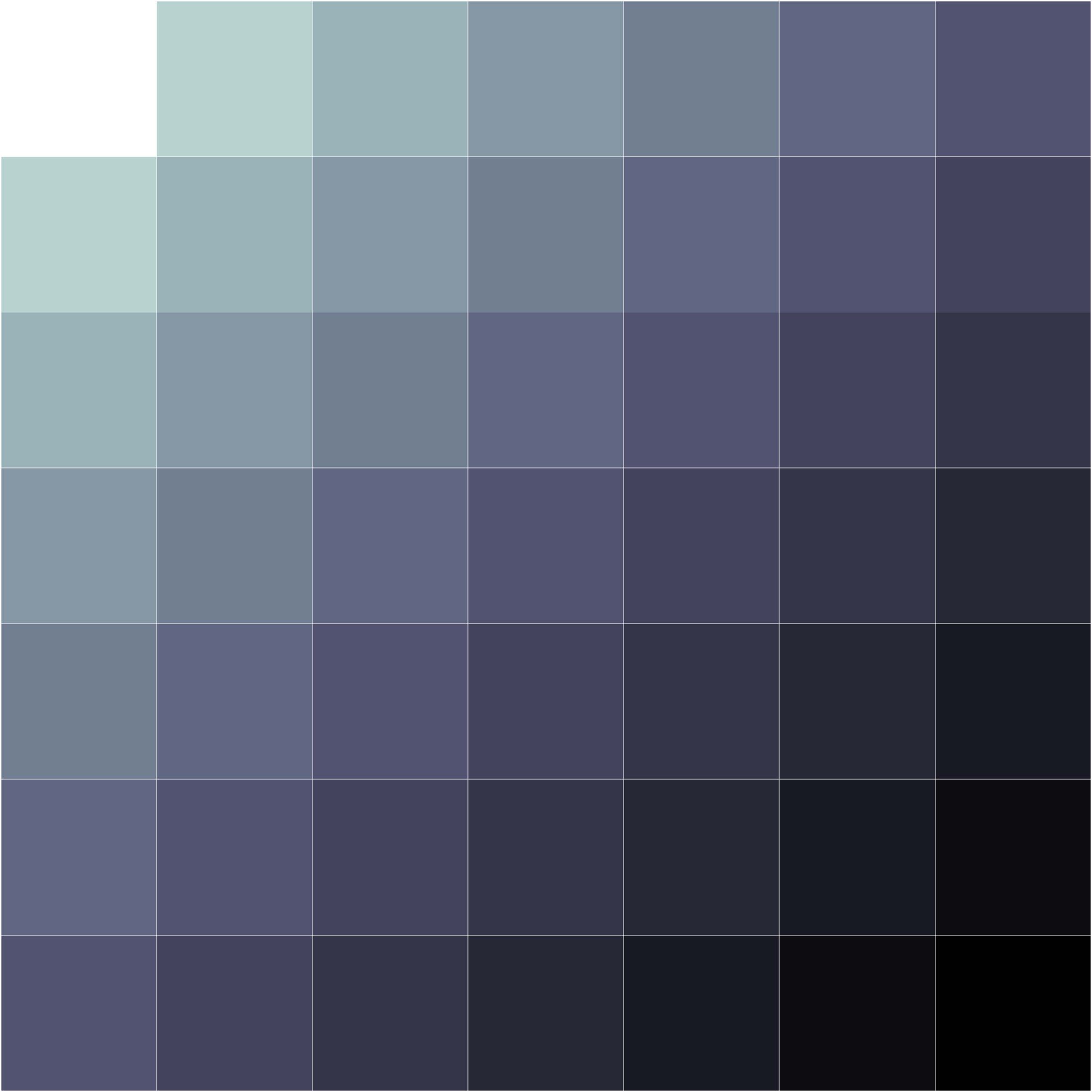}}
          \caption{$\tiny{\mat{H}^{(1)}_n}$}
       \label{fig:Poly1D}
       \end{subfigure}
       \begin{subfigure}[2D]{0.16\textwidth}
           \centering
           \stackunder[5pt]{2D}{\includegraphics[width=\textwidth]{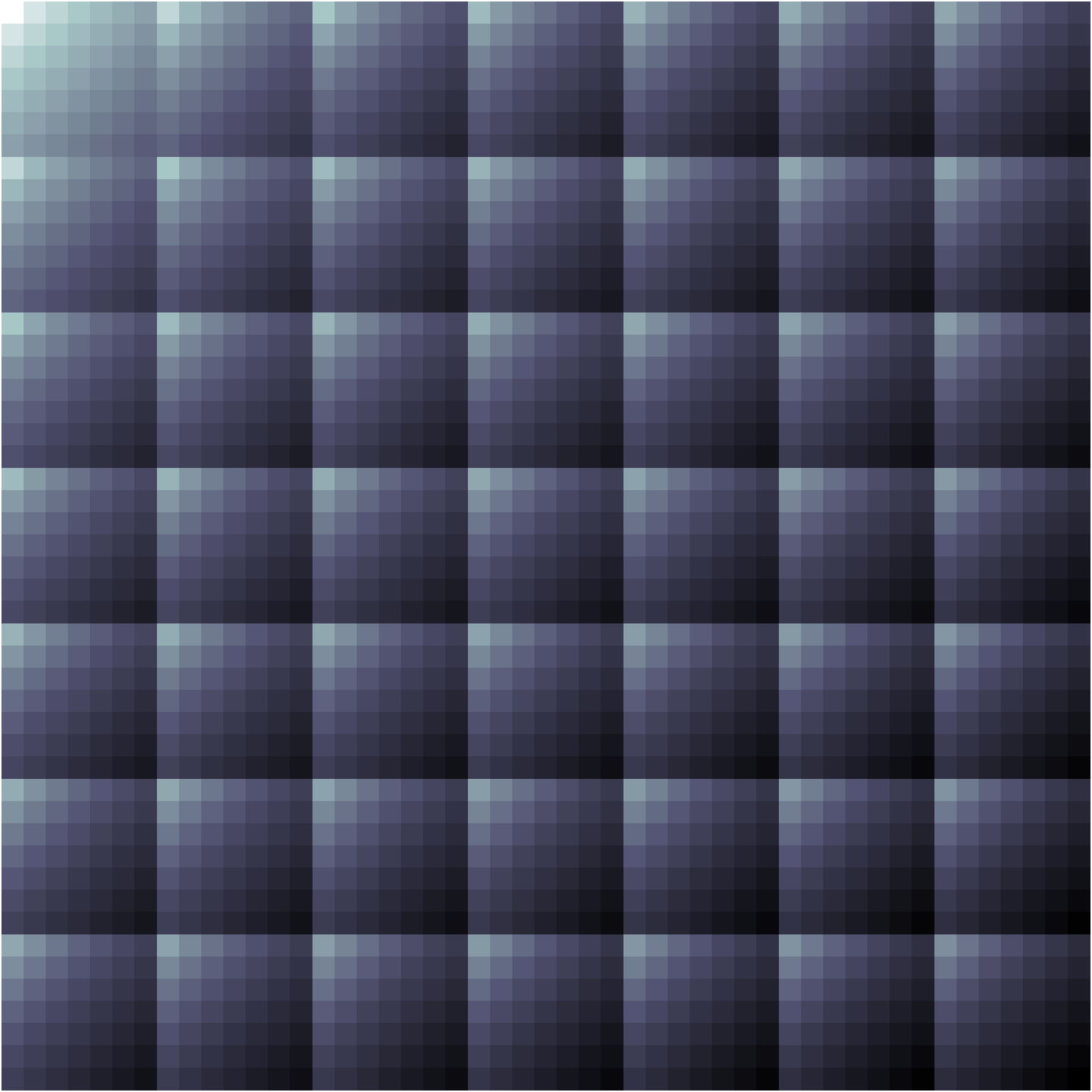}}
           \caption{ $\tiny{\mat{H}_n^{(1)}\otimes \mat{H}_n^{(2)}}$}
           \label{fig:Poly2D}
       \end{subfigure}
       \begin{subfigure}[3D]{0.16\textwidth}
           \centering
           \stackunder[5pt]{3D}{\includegraphics[width=\textwidth]{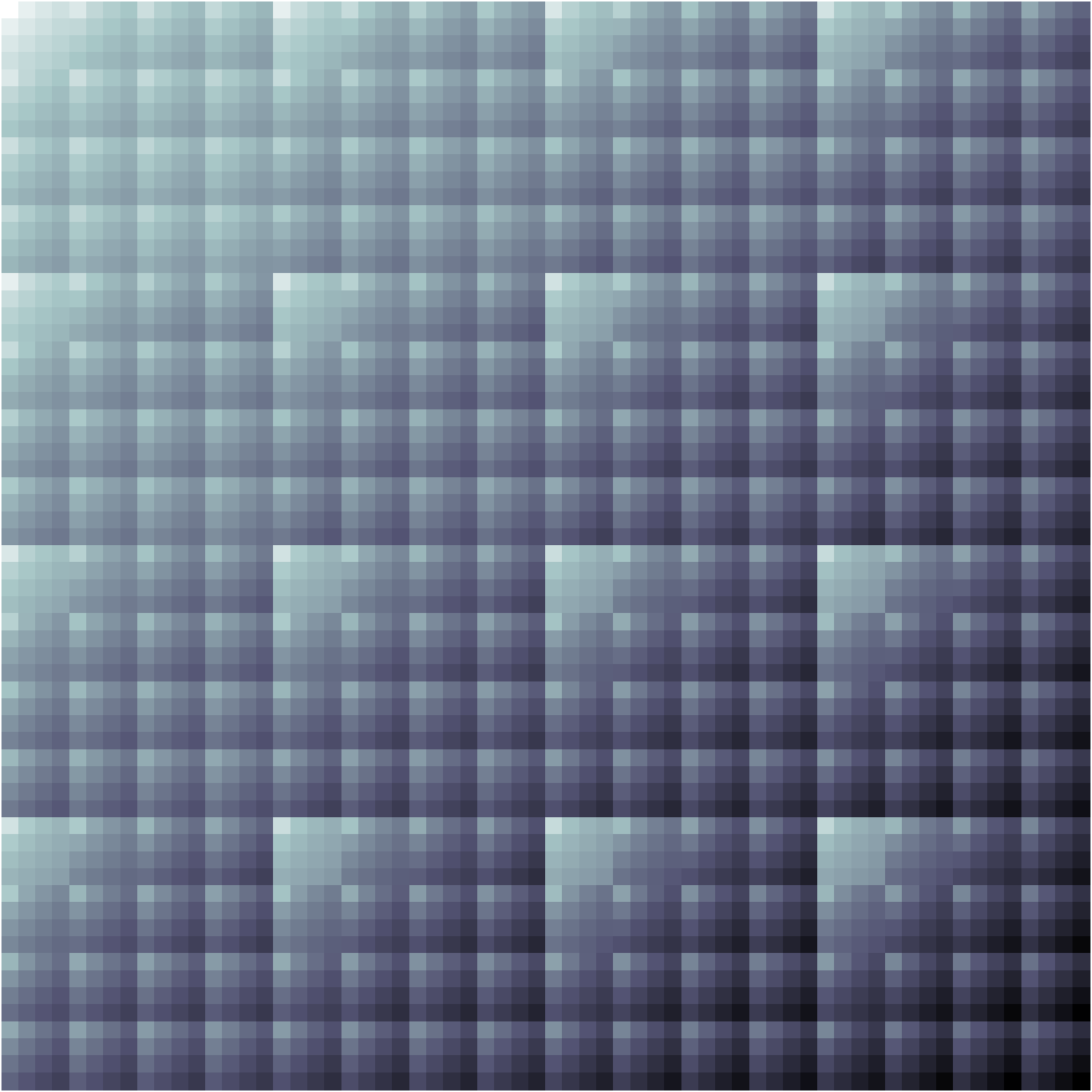}}
           \caption{$\tiny{\mat{H}^{(1)}_n\!\otimes\! \mat{H}^{(2)}_n\!\otimes\! \mat{H}^{(3)}_n}$}
           \label{fig:Poly3D}
       \end{subfigure}
      \caption{The precision matrix for polynomial basis functions has a nested Hankel structure. The visualization of the matrix is proportionally darker as the logarithm of each entry increases. The matrices are computed as the sum of all entries $\mat{H}_n$ for $n=\{1,\hdots, N\}$, where the expression for $\mat{H}_n$ is given below each matrix.} 
      \label{fig:MonomialFunctions}
\end{figure}

In this paper, we reduce this complexity to $\mathcal{O}(NM)$ with \emph{no additional approximations} (see \cref{fig:timing}), by exploiting structural properties of the precision matrix. 
We remark that these structural properties arise from the \glspl{bf} that are used to approximate the kernel, \emph{not} the kernel itself.
Our contributions are as follows.
\begin{itemize}
    \item We show that the \gls{hgp} precision matrix using basis functions defined on a \mbox{(hyper-)cubical} domain can be split in a sum of multilevel block-Hankel and a multilevel block-Toeplitz matrix (\cref{fig:MonomialFunctions}), where each summand only has $\mathcal{O}(M)$ unique entries instead of $\mathcal{O}(M^2)$. 
    This allows us to determine the elements of the full precision matrix at $\mathcal{O}(NM)$ instead of $\mathcal{O}(NM^2)$.
    \item This method does not only hold for \gls{hgp} basis functions. We provide sufficient conditions on the \glspl{bf} for this reduction in computational complexity to hold.
    Examples of other \glspl{bf} where the conditions hold, and that therefore can be sped up using the same technique are polynomial, (complex) exponential and (co)sinusoidal basis functions. 
    We, therefore, enable the speed-up of other \gls{bf}-based \gls{gp} approximations such as variational Fourier features \citep{hensmanVariational2017}.
\end{itemize}
In the experiments, we demonstrate in practice that our approach lowers the computational complexity of the \gls{hgp} by an order of magnitude on simulated and real data.

\section{Background}\label{sec:background}

A \gls{gp} is a collection of random variables, any finite number of which have a joint Gaussian distribution \citep{rasmussen2005gaussian}.
We denote a zero--mean \gls{gp} by ${f \sim \mathcal{GP}(0, \kappa(\cdot, \cdot))}$, where ${\kappa(\mat{x},\mat{x}'):\mathbb{R}^D \times \mathbb{R}^D \to \mathbb{R}}$ is the kernel, representing the covariance between inputs $\mat{x}$ and $\mat{x}'$. Given a dataset of input--output pairs $\{(\mat{x}_n,y_n)\}_{n=1}^N$, \glspl{gp} are used for non-parametric regression and classification by coupling the latent functions $f$ with observations through a likelihood model $p(\mat{y}\mid f) = \prod_{i=1}^N p(y_i \mid f(\mat{x}_i))$.

For notational simplicity, we will in the following focus on \gls{gp} models with a Gaussian (conjugate) likelihood, $y_i \sim \mathcal{N}(f(\mat{x}_i), \sigma^2)$. The posterior \gls{gp}, $\mathcal{GP}(\mu_\star(\cdot), \Sigma_\star(\cdot,\cdot))$ can be written down in closed form by 
\begin{subequations}\label{eq:gpposteriorpredictive}
    \begin{align}
        \mu_\star(\mat{x_\star}) &= \mat{k}_\star\trans(\mat{K} + \sigma^2\mat{I})^{-1}\mat{y},\\ 
    \Sigma_\star(\mat{x_\star},\mat{{x}_\star '})  &={\mat{k}(\mat{x_\star},\mat{x_\star}')} - \mat{k}_\star\trans(\mat{K} + \sigma^2\mat{I})^{-1}\mat{k}_{\star'},
    \end{align}
\end{subequations}
where $\mat{K}\in  \mathbb{R}^{N\times N}$ and $\mat{k}_\star\in\mathbb{R}^{N}$ are defined element-wise as $\mat{K}_{i,j}\isdef \kappa(\mat{x}_i, \mat{x}_j)$ and $[\mat{k}_\star]_{i} \isdef \kappa(\mat{x}_i, \mat{x}^\star)$ for $i,j\in 1,2,\ldots,N$, and $\mat{k}_{\star'}$ is defined similarly.
Observations are collected into $\mat{y}\in\mathbb{R}^N$. 
Due to the inverse of $(\mat{K}+\sigma^2\mat{I})$ in the posterior mean and covariance, the computational cost of a standard \gls{gp} scales as $\mathcal{O}(N^3)$, which hinders applicability to large datasets.

\subsection{Basis Function Approximations}
The prevailing approach in literature to circumvent the $\mathcal{O}(N^3)$ computational bottleneck is to approximate the \gls{gp} with a sparse approximation, using a finite number of either inducing points or \glsxtrfullpl{bf} \citep[e.g.,][]{rasmussen2005gaussian,quinonero-candelaUnifyingViewSparse2005,hensmanVariational2017}.
The \gls{bf} representation is commonly motivated by the approximation
\begin{equation}
    \kappa(\mat{x},\mat{x}') \approx \mat{\phi}(\mat{x})\trans\mat{\Lambda}~\mat{\phi}(\mat{x}').
\end{equation}
We use the notation from \citet{solinHilbertSpaceMethods2020} to align with the next section. Here, $\mat{\phi}(\cdot):\mathbb{R^D}\to \mathbb{R}^{M}$ are the \glspl{bf}, $\mat{\phi}(\cdot)\isdef [\phi_1(\cdot), \phi_2(\cdot),\dots,\phi_M(\cdot)]\trans$.
Further, $\mat{\Lambda}\in\mathbb{R}^{M\times M}$ are the corresponding \gls{bf} weights.
Combining this approximation with the posterior \gls{gp}, \cref{eq:gpposteriorpredictive}, and applying the Woodbury matrix inversion lemma yields \begin{subequations}\label{eq:bfposteriorpredictive}
    \begin{align}
         \mu_{\star}(\mat{x_\star}) &=\mat{\phi}(\mat{x_\star})\trans \left( \mat{\Phi}\trans\mat{\Phi} + \sigma^2\mat{\Lambda}^{-1} \right)^{-1}\mat{\Phi}\trans \mat{y},\\
         \Sigma_{\star}(\mat{x_\star},\mat{x_\star}')&=\sigma^2 \mat{\phi}(\mat{x_\star})\trans\left(\mat{\Phi}\trans\mat{\Phi} + \sigma^2\mat{\Lambda}^{-1} \right)^{-1}{\mat{\phi}(\mat{x_\star'})}.
    \end{align}
\end{subequations}
Here, $\mat{\Phi} \in \mathbb{R}^{N\times M}$ is commonly referred to as the regressor matrix and is defined as $\mat{\Phi}_{i,:} \isdef \mat{\phi}(\mat{x}_i)\trans$.
Further, $\mat{\Phi}\trans\mat{\Phi}\in\mathbb{R}^{M\times M}$ is the \emph{precision matrix} which is a central component of the following section. Computing this precision matrix requires $\mathcal{O}(NM^2)$ operations. With this approximation, computing the posterior mean and covariance in \cref{eq:bfposteriorpredictive} requires instantiating and inverting $\left(\mat{\Phi}\trans\mat{\Phi} + \sigma^2\mat{\Lambda}^{-1} \right)^{-1}$ which can be performed with $\mathcal{O}(NM^2+M^3)$ operations. If we have more samples than \glspl{bf} (which is the required condition for \glspl{bf} to give computational savings), i.e., $N\geq M$, the overall computational complexity is then of $\mathcal{O}(NM^2)$, i.e., the inversion costs are negligible as compared to the cost of computing the precision matrix.

\section{Methods}\label{sec:method}

Our main findings are in the form of two theorems (\cref{thm:2md,thm:3md}). These theorems prescribe the necessary conditions that \gls{bf} expansions need to fulfill to be able to reduce the computational complexity of computing the precision matrix $\mat{\Phi}\trans\mat{\Phi}$ from $\mathcal{O}(NM^2)$ to $\mathcal{O}(NM)$, applicable to multiple previous works that rely on parametric basis functions \citep[incl.][]{lazaro-gredilla_sparse_spectrum, hensmanVariational2017, solinHilbertSpaceMethods2020, Tompkins_Ramos_2018, pmlr-v119-dutordoir20a}. Since our contribution reduces the cost of computing the precision matrix to $\mathcal{O}(NM)$, it follows that the computational complexity of computing the posterior mean or covariance is also of $\mathcal{O}(NM)$.
Further, both of the theorems 
reduce the memory scaling from $\mathcal{O}(M^2)$ to $\mathcal{O}(M)$.
Note that these reductions are \emph{without} approximations, only relying on the structural properties of the considered models.

In the following, we assume that the kernel is a tensor product kernel \citep{rasmussen2005gaussian}, i.e., $\kappa(x,x')=\prod_{d=1}^D\kappa^{(d)}(x^{(d)},{x}^{(d)'})$, where $\kappa^{(d)}(\cdot,\cdot)$ is the kernel along the $d$\textsuperscript{th} dimension.
Then, if each component of the kernel is approximated with $m_d$ \gls{bf}s such that
 \begin{equation}
    \kappa^{(d)}(x^{(d)},{x^{(d)}}') \approx {\mat{\phi}^{(d)}(x^{(d)})}\trans \mat{\Lambda}^{(d)} \ \mat{\phi}^{(d)}({x^{(d)}}'),
 \end{equation}
where $\mat{\phi}^{(d)}(\cdot):\mathbb{R}\to\mathbb{R}^{m_d}$ are the \glspl{bf} $[\phi^{(d)}_{1},\phi^{(d)}_{2},\hdots,\phi^{(d)}_{m_d}]\trans$ along the $d$\textsuperscript{th} dimension and $\mat{\Lambda}^{(d)}\in\mathbb{R}^{m_d\times m_d}$ contains the associated weights.
The full kernel can then be approximated as
 \begin{equation}\label{eq:fullproductkernel}\textstyle
     \kappa(\mat{x},\mat{x}') \approx \textstyle\prod_{d=1}^D {\mat{\phi}^{(d)}({x}^{(d)})}\trans \mat{\Lambda}^{(d)} \ \mat{\phi}^{(d)}({{x}^{(d)}}'),
 \end{equation}
where in this case ${\mat{\phi}(\cdot):\mathbb{R}^{D}\rightarrow\mathbb{R}^{M}}$ and ${\mat{\Lambda}\in\mathbb{R}^{M\times M}}$ are
 \begin{subequations}
     \begin{align}
         \mat{\phi}(\mat{x}) &= \otimes_{d=1}^D \mat{\phi}^{(d)}(x^{(d)}), \label{eq:kroenecker_BFs}\\
         \mat{\Lambda} &= \otimes_{d=1}^D \mat{\Lambda}^{(d)}.
     \end{align}
 \end{subequations}
Here, $M \isdef \prod_{d=1}^D m_d$ is the total number of \glspl{bf}.
Given this decomposition, the precision matrix can be expressed as\looseness-1
\begin{equation}\label{eq:productprecisionmatrix}
    \mat{\Phi}\trans\mat{\Phi} =\textstyle\sum_{n=1}^N \mat{\phi}(\mat{x}_n) \mat{\phi}(\mat{x}_n)\trans
    =\textstyle\sum_{n=1}^N\otimes_{d=1}^D \mat{\phi}^{(d)}(x^{(d)}_n) \left[\mat{\phi}^{(d)}(x^{(d)}_n)\right]\trans.
\end{equation}
This decomposition of the precision matrix is key in the following and we will primarily study the individual products $\mat{\phi}^{(d)}(x^{(d)}_n) [\mat{\phi}^{(d)}(x^{(d)}_n)]\trans$ where certain structure may appear that is exploitable to our benefit.
To provide some intuition, we consider the precision matrix for polynomial \glspl{bf} in 1D, 2D, and 3D (see \cref{fig:MonomialFunctions}).
The 1D case (left) has a \emph{Hankel} structure and the 2D (middle) and 3D (right) cases have 2-level and 3-level \emph{block Hankel} structure, respectively.
It is these types of structures that allow a reduction in complexity \emph{without} approximations.
Next, we provide clear technical definitions of the matrix structures and then proceed to state our main findings.
All of the discussed matrix structures are also visually explained in \cref{tab:Hankel_Toeplitz_structures} in \cref{app:Hankel_Toeplitz_structures}.

\subsection{Hankel and Toeplitz Matrices}
\begin{figure}[t]
    \centering
    \begin{equation*}
\underbrace{\includegraphics[valign=c,width=0.13\textwidth]{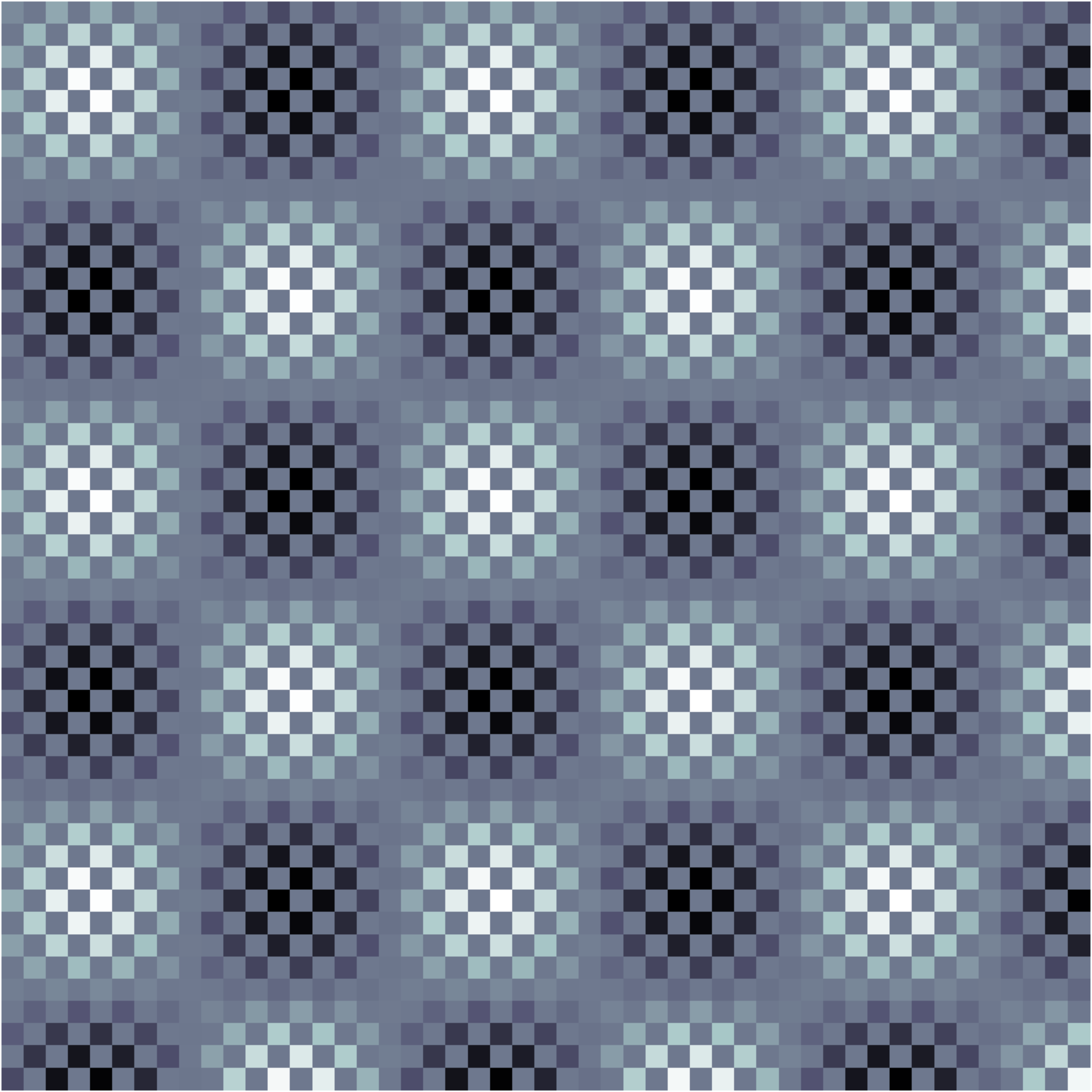}}_{\mat{\phi}^{(d)}{\mat{\phi}^{(d)}}\trans} \: = \: \underbrace{\includegraphics[valign=c,width=0.13\textwidth]{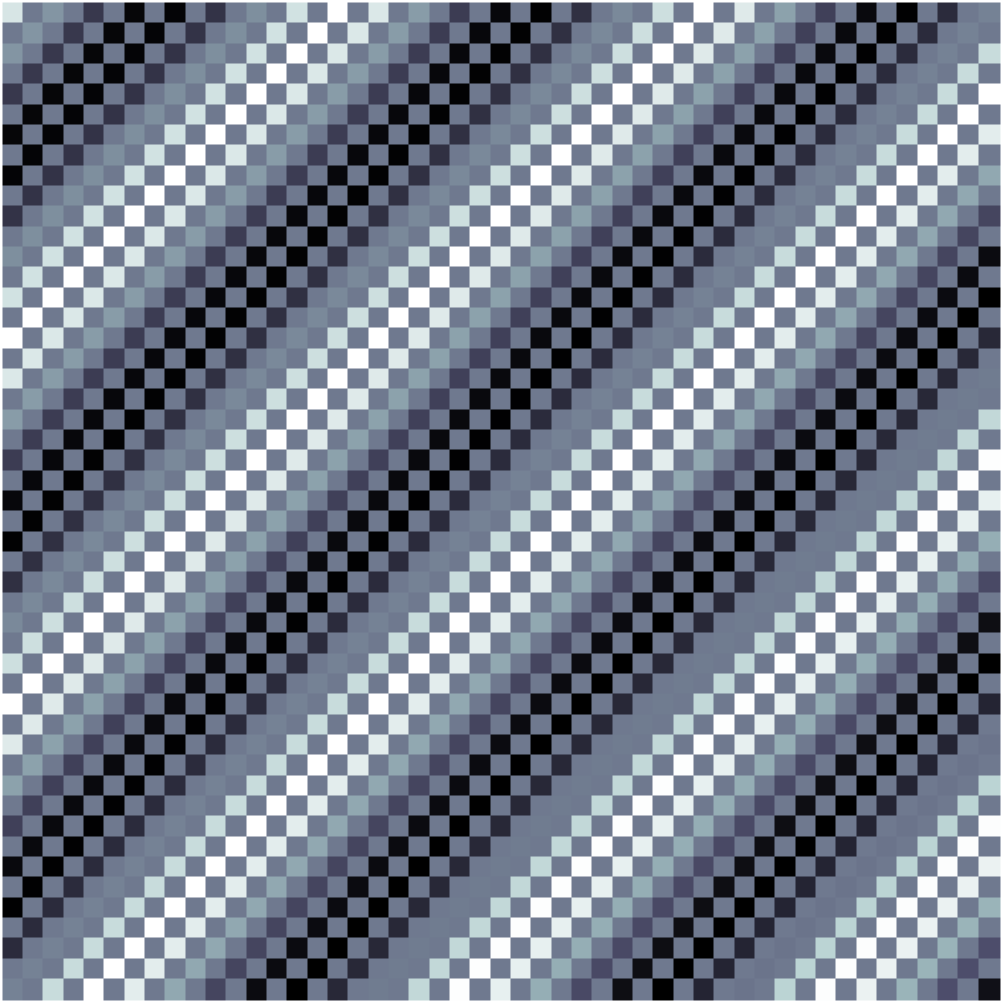}}_{\sum\limits_{n}\mat{H}^{(1)}} + \underbrace{\includegraphics[valign=c,width=0.13\textwidth]{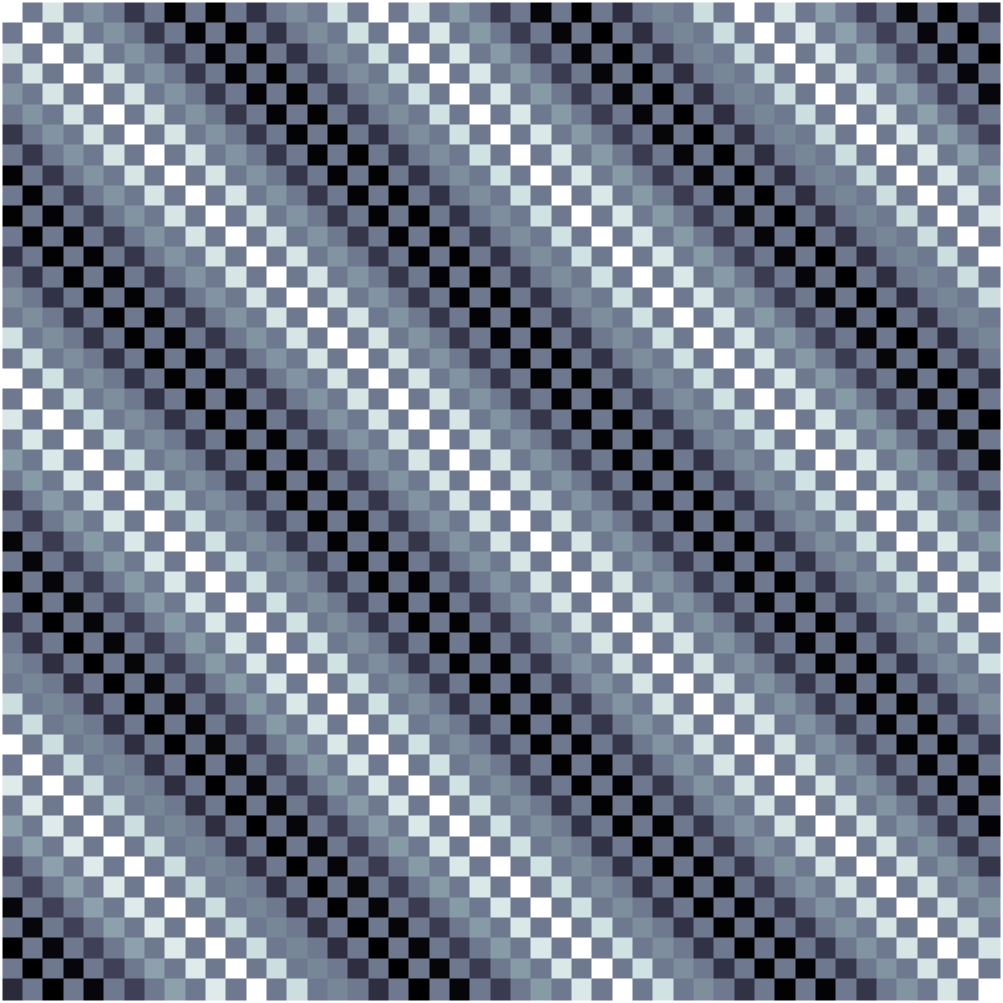}}_{\sum\limits_{n} \mat{T}^{(1)}}\:.
\end{equation*}
\vskip -.1in
    \caption{The precision matrix for sinusoidal basis functions in one dimension has neither Hankel nor Toeplitz structure. However, it can be decomposed into a sum of two matrices, where one has a Hankel structure, and one has Toeplitz structure. Here, 49 \glspl{bf} are placed along one dimension.}
    \label{fig:PrecisionMatrixHS1D}
\end{figure}

An $m\times m$ matrix $\mat{H}$ has Hankel structure iff it can be expressed using a vector $ \mat{\gamma} \in \mathbb{R}^{(2m-1)}$ containing all the unique entries, according to
\begin{equation}
   \mat{H}=\begin{bmatrix}
\mat{\gamma}_1 & \mat{\gamma}_2 & \hdots & \mat{\gamma}_m\\ 
\mat{\gamma}_2 & \mat{\gamma}_3 & \hdots & \mat{\gamma}_{m+1}\\ 
\vdots & \vdots & \ddots & \vdots \\ 
\mat{\gamma}_{m} & \mat{\gamma}_{m+1} & \hdots & \mat{\gamma}_{2m-1}
\end{bmatrix}. 
\end{equation}
In other words, each element of the Hankel matrix on the $i$\textsuperscript{th} row and the $j$\textsuperscript{th} column is given by $\mat{\gamma}_{i+j-1}$.
Similarly, an $m\times m$ matrix has Toeplitz structure iff it can be expressed using a vector $\mat{\gamma} \in \mathbb{R}^{(2m-1)}$ such that each element on the $i$\textsuperscript{th} row and the $j$\textsuperscript{th} column is given by $\mat{\gamma}_{i-j+m}$.
See \cref{fig:PrecisionMatrixHS1D} for an example of what a Hankel and a Toeplitz matrix visually looks like.

We define a matrix $\mat{H}^{(D)}$ as a $D$-level block Hankel matrix if it can be expressed as
\begin{equation}
    \mat{H}^{(D)}=\begin{bmatrix}
\mat{H}^{(D-1)}_{1} & \mat{H}_{2}^{(D-1)} & \hdots & \mat{H}_{m_D}^{(D-1)}\\ 
\mat{H}_{2}^{(D-1)} & \mat{H}_{3}^{(D-1)} & \hdots & \mat{H}_{m_D+1}^{(D-1)}\\ 
\vdots & \vdots & \ddots & \vdots \\ 
\mat{H}_{m_D}^{(D-1)} & \mat{H}_{m_D+1}^{(D-1)} & \hdots & \mat{H}_{2m_D-1}^{(D-1)}
\end{bmatrix},
\end{equation}
where $\mat{H}_{k}^{(D-1)}$ is a block Hankel matrix, and $\mat{H}_{k}^{(1)}$ is a simple Hankel matrix. 
Each submatrix $\mat{H}_k^{(d)}$ for $d\in\{1,\hdots, D\}$ can be indexed by row and column $i_d,j_d\in\{1,\dots,m_d\}$ which again yields a submatrix $\mat{H}^{(d-1)}_{i_d+j_d-1}$ which can be indexed in the same manner.
Therefore, an equivalent definition of a block Hankel matrix is that each individual entry of $\mat{H}^{(D)}$, given by a set of indices $i_1, j_1,\dots,i_D,j_D$, is given by the entry $\mat{\gamma}_{k_1,\hdots,k_D}$ of a tensor 
$\mat{\gamma}\in\mathbb{R}^{K_1\times K_2 \times \hdots \times K_D}$, where $k_d=i_d+j_d-1$, and $K_d=2m_d-1$.
Similarly, $\mat{T}^{(D-1)}$ is a block Toeplitz matrix if each entry $i_1,j_1,\hdots,i_D,j_D$ is given by the entry $\mat{\gamma}_{k_1,\hdots,k_D}$ of a tensor $\mat{\gamma}\in\mathbb{R}^{K_1\times K_2 \times \hdots \times K_D}$, where $k_d=m_d+i_d-j_d$, and $K_d=2m_d-1$.

\subsection{Block Hankel--Toeplitz Matrices}\label{subsec:blockhankeltoeplitz}
In addition to block Hankel and block Toeplitz structures, we require a slightly more general but highly related structure.
We call this structure block Hankel--Toeplitz structure and define it as follows.
A matrix $\mat{G}^{(D)}$ has a $D$-level block Hankel--Toeplitz structure, if the matrix is defined either as
\begin{equation}\label{eq:G_Hankel_structure}
    \mat{G}^{(D)}=\begin{bmatrix}
\mat{G}^{(D-1)}_{1} & \mat{G}_{2}^{(D-1)} & \hdots & \mat{G}_{m_D}^{(D-1)}\\ 
\mat{G}_{2}^{(D-1)} & \mat{G}_{3}^{(D-1)} & \hdots & \mat{G}_{m_D+1}^{(D-1)}\\ 
\vdots & \vdots & \ddots & \vdots \\ 
\mat{G}_{m_D}^{(D-1)} & \mat{G}_{m_D+1}^{(D-1)} & \hdots & \mat{G}_{2m_D-1}^{(D-1)}
\end{bmatrix},
\end{equation}
if level $D$ is Hankel, or as 
\begin{equation}\label{eq:G_Toeplitz_structure}
    \mat{G}^{(D)}=\begin{bmatrix}
\mat{G}_{m_D}^{(D-1)} & \hdots & \mat{G}_{2}^{(D-1)} & \mat{G}_{1}^{(D-1)}\\ 
\mat{G}_{m_D+1}^{(D-1)} & \hdots & \mat{G}_{3}^{(D-1)} & \mat{G}_{2}^{(D-1)}\\ 
\vdots & \iddots & \vdots & \vdots\\ 
\mat{G}^{(D-1)}_{2m_D-1} & \hdots  & \mat{G}_{m_D+1}^{(D-1)} & \mat{G}_{m_D}^{(D-1)}
\end{bmatrix},
\end{equation}
if level $D$ is Toeplitz.
Further, $\mat{G}^{(D-1)}_{j}$ are $D-1$ level block Hankel--Toeplitz matrices if $D-1>2$, and a simple Hankel or Toeplitz matrix if $D-1=1$. 
Each submatrix $\mat{G}_k^{(d)}$ for $d\in\{1,\hdots, D\}$ can be indexed by row and column $i_d,j_d\in\{1,\dots,m_d\}$ which again yields a submatrix defined either as $\mat{G}^{(d-1)}_{i_d+j_d-1}$ or $\mat{G}^{(d-1)}_{m_d+i_d-j_d}$ (depending on whether the $d$\textsuperscript{th}-level has Hankel structure as in \cref{eq:G_Hankel_structure} or Toeplitz structure as in \cref{eq:G_Toeplitz_structure}). 
Each entry in a block Hankel--Toeplitz matrix can also be expressed by the entry $\mat{\gamma}_{k_1,k_2,\hdots,k_D}$ of a tensor $\mat{\gamma}\in\mathbb{R}^{K_1\times K_2\times\hdots\times K_D}$, where $K_d=2m_d-1$, and
\begin{equation}\label{eq:k_d}
k_d = \begin{cases}
    i_d+j_d-1, & \text{if level $d$ is Hankel}\\ 
m_d+i_d-j_d, & \text{if level $d$ is Toeplitz}.
\end{cases}
\end{equation}
A crucial property of the block Hankel--Toeplitz structure is the preservation of structure under addition.
Assume that $\mat{A}$ and $\mat{B}$ are two block Hankel--Toeplitz matrices and that they are structurally identical, in the sense that they have the same number of levels, the same number of entries in each block, and each level shares either Toeplitz or Hankel properties.
Then, let each entry of $\mat{A}$ and $\mat{B}$ be given by $\alpha_{k_1,k_2,\hdots,k_D}$ for a tensor $\alpha\in\mathbb{R}^{m_1,\hdots,m_D}$ 
and $\beta_{k_1,k_2,\hdots,k_D}$ for $\beta\in\mathbb{R}^{m_1,\hdots,m_D}$, respectively, with $k_d$ defined in \cref{eq:unique_entries_block_hankel}.
Each entry in $\mat{A}+\mat{B}$ is then given by the sum of the entries $\alpha_{k_1,\hdots,k_D}+\beta_{k_1,\hdots,k_D}$.
Thus, the sum of two block Hankel--Toeplitz matrices with identical structure is also a block Hankel--Toeplitz matrix. 
By associativity of matrix addition, a sum $\sum_{n=1}^{N}\mat{G}_n$ of $N$ Hankel--Toeplitz matrices $\{\mat{G}_1,\hdots,\mat{G}_N\}$ with identical structure is therefore itself a Hankel--Toeplitz matrix.

\subsection{Kronecker Products of Hankel--Toeplitz Matrices and Block Hankel--Toeplitz Matrices}\label{subsec:kroneckerstructure}
A special case of a class of matrices $\mat{G}$ which has block Hankel--Toeplitz structure are Kronecker products of $D$ Hankel and Toeplitz matrices $\{ \mat{G}^{(1)}, \hdots, \mat{G}^{(D)}\}$, i.e., 
\begin{equation}
    \mat{G}=\bigotimes_{d=1}^D \mat{G}^{(d)} \isdef \mat{G}^{(1)}\otimes \mat{G}^{(2)}\otimes\dots\otimes\mat{G}^{(D)}.
\end{equation}
An equivalent definition of the Kronecker product gives each entry on the $i$\textsuperscript{th} row and $j$\textsuperscript{th} column in the block Hankel--Toeplitz matrix $\mat{G}$ as an expression of the entries on the $i_d$\textsuperscript{th} row and $j_d$\textsuperscript{th} column of each matrix $\mat{G}^{(d)}$ according to
\begin{equation}
    \mat{G}_{i,j}=\textstyle\prod_{d=1}^D \mat{G}_{i_d,j_d}^{(d)}.
\end{equation}
Note that there is a one-to-one map between each index $i,j$ and the index sets $\{i_1,\hdots,i_D\}$ and $\{j_1,\hdots,j_D\}$. 
As each matrix $\mat{G}^{(d)}$ has Hankel or Toeplitz structure, the entries can equivalently be defined by a vector $\mat{\gamma}^{(d)}$ with $2m_d-1$ entries. Each entry $\mat{G}_{i,j}$ is therefore given by
\begin{equation}\label{eq:unique_entries_block_hankel}    \mat{G}_{i,j}=\textstyle\prod_{d=1}^D \mat{\gamma}^{(d)}_{k_d}=\mat{\gamma}_{k_1,\hdots,k_D},
\end{equation}
where $k_d$ is defined in~\cref{eq:k_d}, and where $\mat{\gamma}:=\bigotimes_{d=1}^D \mat{\gamma}^{(d)}$ is a rank-1 tensor with $\prod_{d=1}^D(2m_d-1)$ elements. 

\subsection{Main Results}
\label{sec:main}
We are now ready to state our main findings.
The following two theorems rely on the product decomposition \cref{eq:fullproductkernel} and study each dimension $d$ separately, as is evidently possible from \cref{eq:productprecisionmatrix}. Our first theorem generalizes a result by~\citet{greengard2023equispaced} regarding complex exponential basis functions. Our first theorem establishes that if the product $\mat{\phi}^{(d)}(x^{(d)}_n) [\mat{\phi}^{(d)}(x^{(d)}_n)]\trans$ has Hankel or Toeplitz structure, the resulting precision matrix only has $\prod_{d=1}^D (2m_d-1)$ unique entries, reducing the computational complexity of instantiating it from $\mathcal{O}(NM^2)$ to $\mathcal{O}(NM)$. 
We formalize this in the following theorem.

\begin{theorem}\label{thm:2md}
    If the matrix 
    \begin{equation}
        \mat{G}^{(d)}(x_n^{(d)})\isdef\mat{\phi}^{(d)}(x^{(d)}) \big[\mat{\phi}^{(d)}(x^{(d)})\big]\trans,
    \end{equation}
    is a Hankel or Toeplitz matrix for all $x^{(d)}\in\mathbb{R}$ along each dimension $d$, the information matrix $\mat{\Phi}\trans\mat{\Phi}$ will be a multi-level block Hankel or Toeplitz matrix, and therefore have ${\prod_{d=1}^D(2m_d-1)}$ unique entries.
\end{theorem}

\begin{proof}
    Assume that the matrix $\mat{G}^{(d)}(x_n^{(d)}) \isdef \mat{\phi}^{(d)}(x^{(d)}_n) [\mat{\phi}^{(d)}(x^{(d)}_n)]\trans$ is Hankel or Toeplitz. 
    The precision matrix can then be expressed as
    \begin{equation}  
    \mat{\Phi}\trans\mat{\Phi} =\textstyle\sum_{n=1}^N\otimes_{d=1}^D \mat{\phi}^{(d)}(x^{(d)}_n) \left[\mat{\phi}^{(d)}(x^{(d)}_n)\right]\trans
    =\textstyle\sum_{n=1}^N \otimes_{d=1}^D \mat{G}^{(d)}(x_n^{(d)}),  
    \end{equation}
    where the matrix $\otimes_{d=1}^D \mat{G}^{(d)}(x_n^{(d)})$ is multi-level Hankel or Toeplitz by definition, see \cref{subsec:kroneckerstructure}. 
    Further, the sum of several D-level block Hankel--Toeplitz matrices is itself a $D$-level block Hankel--Toeplitz matrix, see \cref{subsec:blockhankeltoeplitz}.
    Since each matrix $\mat{G}^{(d)}(x_n^{(d)})$ has at most $(2m_d-1)$ unique entries, the matrix $\mat{\Phi}\trans\mat{\Phi} = \sum_{n=1}^N \otimes_{d=1}^D \mat{G}^{(d)}(x_n^{(d)})$ therefore has at most $M=\prod_{d=1}^D (2m_d-1)$ unique entries.
\end{proof}

The preceding theorem holds true for, for instance, polynomial and complex exponential \glspl{bf}, which we establish in \cref{cor:polynomial,cor:complexexponential}.
\begin{corollary}\label{cor:polynomial}
    The precision matrix for polynomial \glspl{bf} defined by
    \begin{equation}
        \phi_{i_d}^{(d)}(x^{(d)})=(x^{(d)})^{i_d-1},
    \end{equation}
    can be represented by a tensor with $\prod_{d=1}^D 2m_d - 1$ entries.
\end{corollary}
\begin{proof}
    See \cref{app:polynomial} for a proof.
\end{proof}
\begin{corollary}\label{cor:complexexponential}
    The precision matrix for complex exponential \glspl{bf} defined by
    \begin{equation}
        \phi_{j}^C(x)=\exp(i\pi j\trans x)=\textstyle\prod_{d=1}^D \exp(i \pi{j_d} x_{d}),
    \end{equation}
    can be represented by a tensor with $\prod_{d=1}^D 2m_d - 1$ entries.
\end{corollary}
\begin{proof}
    See \cref{app:complexexponentialproof} for a proof.
\end{proof}

\begin{figure}[t]
    \centering\large
    \newlength{\blocksize}\setlength{\blocksize}{0.15\textwidth}
    \newlength{\blockspace}\setlength{\blockspace}{4pt}
    \begin{equation*}
  \underbrace{\includegraphics[valign=c,width=\blocksize]{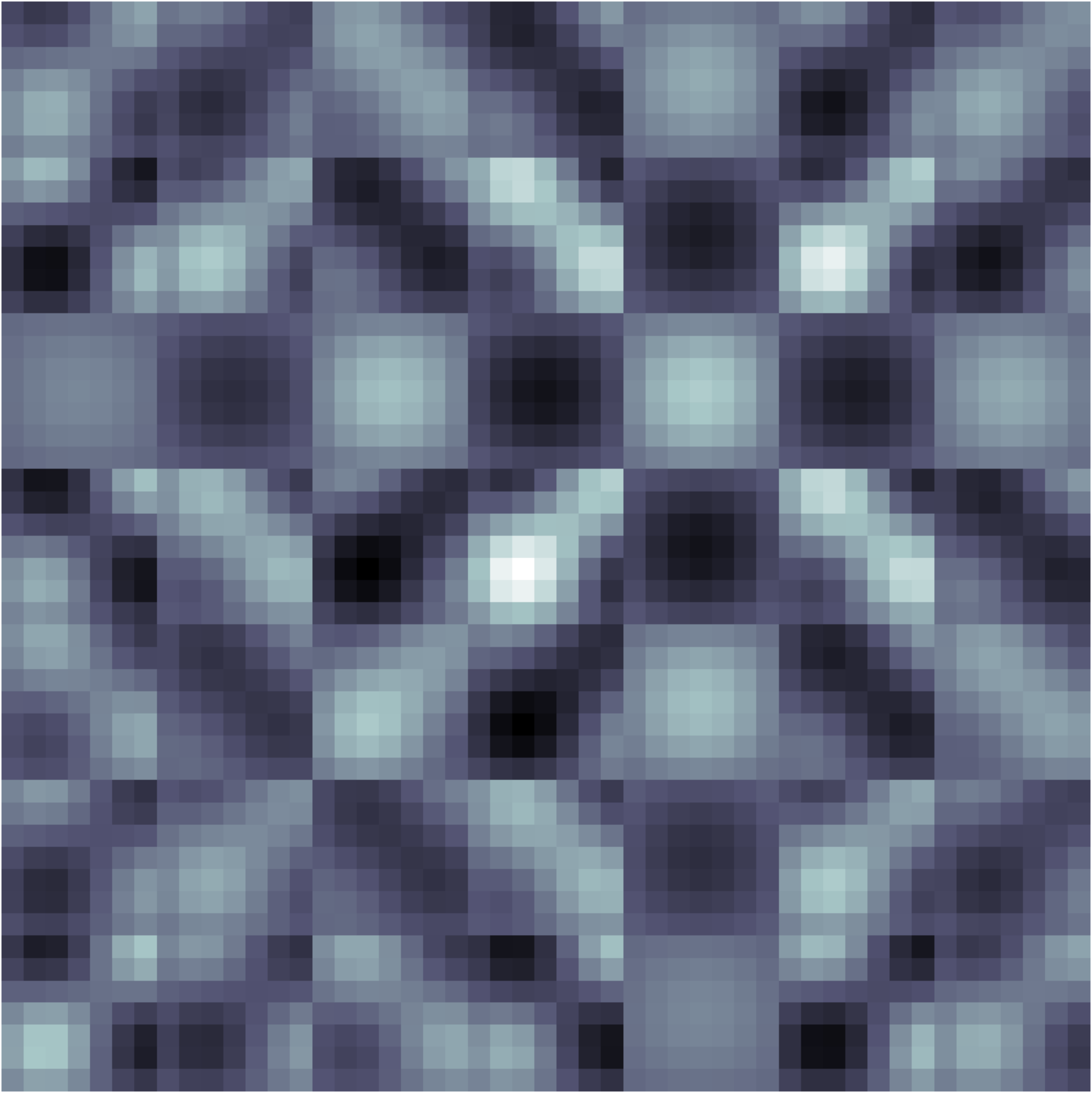}}_{\Phi \trans\Phi}
  \hspace*{\blockspace}\: = \:\hspace*{\blockspace} \underbrace{\includegraphics[valign=c,width=\blocksize]{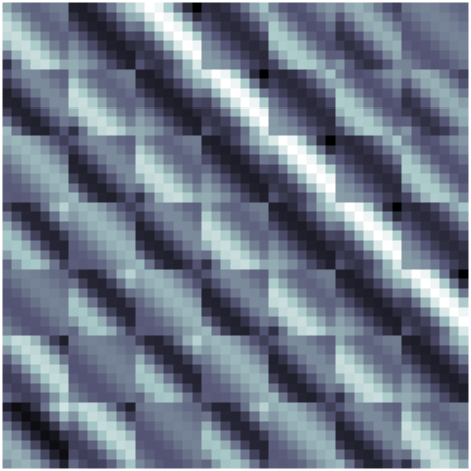}}_{\sum\limits_{n} \mat{H}^{(1)}\otimes \mat{H}^{(2)}}
  \hspace*{\blockspace}+\hspace*{\blockspace}
 \underbrace{\includegraphics[valign=c,width=\blocksize]{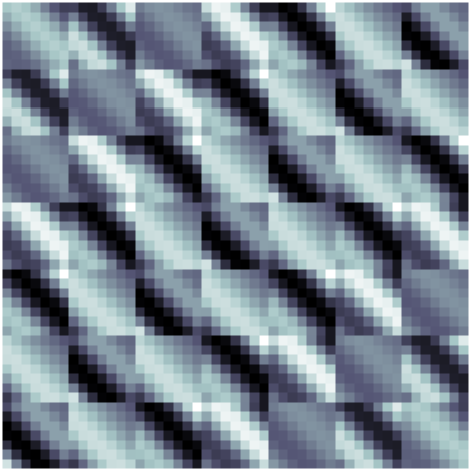}}_{\sum\limits_{n} \mat{T}^{(1)}\otimes \mat{H}^{(2)}} 
  \hspace*{\blockspace}+\hspace*{\blockspace}
\underbrace{\includegraphics[valign=c,width=\blocksize]{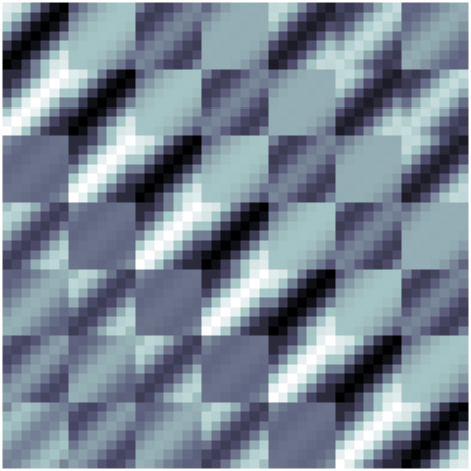}}_{\sum\limits_{n} \mat{H}^{(1)}\otimes \mat{T}^{(2)}} 
  \hspace*{\blockspace}+\hspace*{\blockspace}
 \underbrace{\includegraphics[valign=c,width=\blocksize]{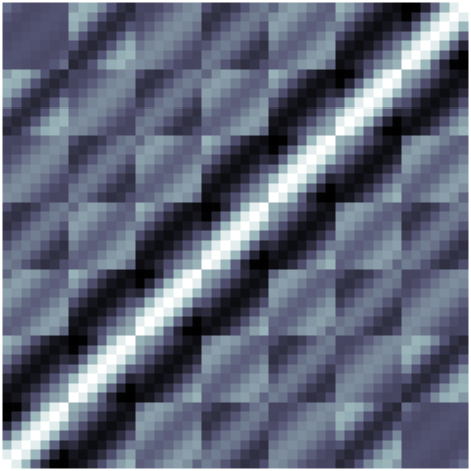}}_{\sum\limits_{n} \mat{T}^{(1)}\otimes \mat{T}^{(2)}}
\end{equation*}
\vskip -.2in
\caption{The precision matrix for sinusoidal \glspl{bf} in two dimensions has neither Hankel nor Toeplitz structure. However, it can be decomposed into $2^D=4$ matrices, which each have block Hankel--Toeplitz structure. Here, 7 \glspl{bf} are placed along each of the two dimensions, giving a total of 49 \glspl{bf}.}
\label{fig:hankelToeplitz}
\end{figure}

For some \glspl{bf}, the structure of the product $\mat{\phi}^{(d)}(x^{(d)}_n) [\mat{\phi}^{(d)}(x^{(d)}_n)]\trans$ is more intricate, but is still of a favorable, exploitable nature.
This is clearly evident from \cref{fig:hankelToeplitz}, where the precision matrix for sinusoidal \glspl{bf} in two dimensions is visualized.
In particular, for some \glspl{bf}, the product is the sum of a Hankel and a Toeplitz matrix, such that the precision matrix only has $\prod_{d=1}^D 3m_d$ unique entries, again reducing the computational cost of computing it to $\mathcal{O}(NM)$.
We formalize this in the following theorem.

\begin{theorem}\label{thm:3md}
If the product $\mat{\phi}^{(d)}(x^{(d)}_n) [\mat{\phi}^{(d)}(x^{(d)}_n)]\trans$ is the sum of a Hankel matrix denoted $\mat{G}^{(d),(1)}(x^{(d)}_n)$ and a Toeplitz matrix $\mat{G}^{(d),(-1)}(x^{(d)}_n)$, and there exists a function $g^{(d)}(k_d,x^{(d)}_n)$ such that $\mat{G}^{(d),(1)}_{i_d,j_d}(x^{(d)}_n)=g(i_d+j_d,x^{(d)}_n)$ and $\mat{G}^{(d),(-1)}_{i_d,j_d}(x^{(d)}_n)=-g(i_d-j_d,x^{(d)}_n)$, all entries in the precision matrix can be represented by a tensor $\mat{\gamma}(k_1,k_2,\hdots,k_D)$ with $\prod_{d=1}^D 3m_d$ entries.
\end{theorem}

\begin{proof}
The precision matrix can in this case be expressed as\looseness-1
\begin{align}
\overbrace{\mat{\Phi}\trans\mat{\Phi}}^{\isdef \mat{C}} &=\textstyle\sum_{n=1}^N\otimes_{d=1}^D \mat{\phi}^{(d)}(x^{(d)}_n) \left[\mat{\phi}^{(d)}(x^{(d)}_n)\right]\trans \nonumber\\
     &=\textstyle\sum_{n=1}^N\otimes_{d=1}^D \left( \mat{G}^{(d),(1)}(x^{(d)}_n) + \mat{G}^{(d),(-1)}(x^{(d)}_n) \right)\nonumber\\
     &=\textstyle\sum_{p=1}^{2^D}\left(\textstyle\prod_{d=1}^D e_{p}^{(d)}  \right) \textstyle\sum_{n=1}^N \otimes_{d=1}^D \mat{G}^{(d),(e_{p}^{(d)})}(x^{(d)}_n),
\end{align}
where $e_p=\{e_{p}^{(1)},\hdots,e_{p}^{(D)}\}\in S^D$ and $S^D=\{1,-1\}^D$ is a set containing $2^D$ elements. 
Each of the $2^D$ matrices $\left(\prod_{d=1}^D e_{p}^{(d)}  \right) \sum_{n=1}^N \otimes_{d=1}^D \mat{G}^{(d),(e_{p}^{(d)})}(x^{(d)}_n)$ is now the Kronecker product between $D$ Hankel or Toeplitz matrices.
The entries of $\mat{C}$ can be expressed element-wise as
\begin{equation}
\mat{C}_{i,j}=\textstyle\sum_{p=1}^{2^D} \left(\textstyle\prod_{d=1}^D e_{p}^{(d)}  \right) \textstyle\sum_{n=1}^N \textstyle\prod_{d=1}^D g(i_d+e_p^{(d)}j_d,x^{(d)}_n).
\end{equation}
If we define a tensor $\mat{\gamma}$ as
\begin{equation}\label{eq:gamma}
\mat{\gamma}_{k_1,\hdots,k_D}=\textstyle\sum_{n=1}^N\textstyle\prod_{d=1}^D g(k_d, x^{(d)}_n)
\end{equation}
for indices $k_d={1-m_d,2-m_d,\hdots,2m_d-1,2m_d}$, each entry of the precision matrix $\mat{C}_{i,j}$ can be expressed as
\begin{equation}
\mat{C}_{i,j}=\textstyle\sum_{p=1}^{2^D}\left(\textstyle\prod_{d=1}^D e_{p}^{(d)}  \right)\mat{\gamma}_{i_1+e_{p}^{(1)}j_1,\hdots,i_D+e_{p}^{(D)}j_D}.
\end{equation}
As each sum $k_d=i_d+e_{p}^{(d)}j_d$ is an integer between $1-m_d$ and $2m_d$, the tensor $\mat{\gamma}$ will have $\prod_{d=1}^D 3m_d$ entries.
\end{proof}

The preceding theorem applies to, for instance, the \glspl{bf} in an \gls{hgp} \citep{solinHilbertSpaceMethods2020} defined on a rectangular domain, formalized in \cref{cor:hgp}.
It further holds for multiple other works using similar \glspl{bf}, formalized in \cref{cor:fourier}.
We remark that for $D=1$, we require $m_d > 3$ for any savings to take effect, whereas for $D>1$, $m_d\geq 2$ suffices.

\begin{corollary}\label{cor:hgp}
    The precision matrix in an \gls{hgp} defined on a rectangular domain $[-L_1, L_1]\times\dots\times [-L_D, L_D]$ can be represented by a tensor with $\prod_{d=1}^D 3m_d$ entries.
\end{corollary}
\begin{proof}
    See \cref{app:hgprectangular} for a proof.
\end{proof}
\begin{corollary}\label{cor:fourier}
    The precision matrix in a \gls{gp} approximated by sinusoidal and cosine \glspl{bf} with frequencies on a grid (such as the regular Fourier features described in~\citet{hensmanVariational2017} and~\citet{wahls_learning_2014}, the Fourier approximations to periodic kernels described in \citet{Tompkins_Ramos_2018}, the quadrature Fourier features described in \citet{Mutn2018EfficientHDQuadratureFourier}, the equispaced-version of sparse spectrum \glspl{bf} described in \citet{lazaro-gredilla_sparse_spectrum}, or the one-dimensional special case of \citet{pmlr-v119-dutordoir20a}) can be represented by a tensor with $\prod_{d=1}^D 3m_d$ entries.
\end{corollary}
\begin{proof}
    See \cref{app:ordinatyFourierFeatures} for a proof.
\end{proof}

\begin{algorithm}[tb]
   \caption{Sketch of an algorithm for Hilbert \gls{gp} learning and inference. The \textcolor{BrickRed}{original approach} by \citet{solinHilbertSpaceMethods2020} in \textcolor{BrickRed}{red}, \textcolor{RoyalBlue}{our proposed approach in blue}.}
   \label{alg:hgp}
\begin{algorithmic}
   \State {\bfseries Input:} Data as input--output pairs $\{(\mat{x}_i,y_i)\}_{i=1}^N$,
   \State \phantom{\bfseries Input:} test inputs $\mat{x}_\star$, number of basis functions $M$
   \State \textcolor{BrickRed}{Compute $\mat{\Phi}\trans\mat{\Phi}$ at cost $\mathcal{O}(NM^2)$} \Comment{\cref{eq:productprecisionmatrix}}
   \State \textcolor{RoyalBlue}{Compute $\mat{\gamma}$ at cost $\mathcal{O}(NM)$} \Comment{\cref{eq:gamma}} 
   \State \textcolor{RoyalBlue}{Construct $\mat{\Phi}\trans\mat{\Phi}$ using $\mat{\gamma}$ at cost $\mathcal{O}(M^2)$} 
   \Repeat
   \State Optimize \gls{mll} w.r.t.\ hyperparameters at cost $\mathcal{O}(M^3)$
   \Until{Convergence}
   \State Perform \gls{gp} inference using the pre-calculated matrices. This entails computing the posterior mean and covariance, at a computational cost of $\mathcal{O}(M^3)$.
\end{algorithmic}
\end{algorithm}

\subsection{Outlook and Practical Use}
Both \cref{thm:2md,thm:3md} reduce the computational complexity of calculating the entries of the precision matrix \cref{eq:productprecisionmatrix} from $\mathcal{O}(NM^2)$ to $\mathcal{O}(NM)$. This enables us to scale the number of \glspl{bf} significantly more than previously, before running into computational, or storage related, problems.
For clarity, the standard \gls{hgp} is given in \cref{alg:hgp} with the original approach in \textcolor{BrickRed}{red} and our proposed approach in \textcolor{RoyalBlue}{blue}.
As compared to the standard (offline) \gls{hgp}, the only change we make is the computation of the precision matrix.

We remark that computing the posterior mean and variance of the \gls{bf} expansion now costs $\mathcal{O}(NM+M^3)$, now possibly dominated by the $M^3$ term.
This cost also appears in the hyperparameter optimization, as we choose to optimize the \gls{mll} as in \citet{solinHilbertSpaceMethods2020}.
Remedies for this cost are out of scope of this paper, but could potentially be reduced through the use of efficient approximate matrix inverses and trace estimators \citep[see, e.g.,][]{daviesEffectiveCG2015}. 
Another consequence of these theorems is that multi-agent systems that collaborate to learn the precision matrix such as~\citet{jang_multi-robot_2020}, ~\citet{viset2023DistributedMagSLAM} or~\citet{pillonetto_distributed_2018} can do this by communicating $\mathcal{O}(M)$ bits instead of $\mathcal{O}(M^2)$ bits.

\section{Experiments}\label{sec:experiments}

We demonstrate the storage and computational savings of our structure exploiting scheme by means of three numerical experiments.
The experiments demonstrate the practical efficiency of our scheme using the \gls{hgp}.
We reiterate that the savings are \emph{without} additional approximations and the posterior is therefore \emph{exactly} equal to that of the standard \gls{hgp}.
Further, as \glspl{hgp} adhere to \cref{thm:3md}, this demonstration is a representative example of the speedups that can be expected using, e.g., regular Fourier features~\citep{hensmanVariational2017}, or \gls{bf} expansions of periodic kernels~\citep{pmlr-v119-dutordoir20a}.
Since \cref{thm:2md} requires computing and storing only $2^DM$ components whereas \cref{thm:3md} requires $3^DM$, our experiments demonstrate a practical upper bound on the storage and computational savings, a ``worst case''. 

\begin{figure}[tb]
\centering\scriptsize
    \begin{subfigure}{.58\textwidth}
    \centering
    \def\arraystretch{.8}
    \setlength{\tabcolsep}{0pt}
    \begin{tikzpicture}
\begin{groupplot}[
    group style={group size=3 by 1, 
    horizontal sep=.25cm,
    y descriptions at=edge left},
    width=.27\textwidth,
    height=.27\textwidth,
    at={(0cm,0cm)},
    scale only axis,
    point meta min=-2,
    point meta max=6,
    axis on top,
    xmin=7.438,
    xmax=7.678,
    xtick={7.45,  7.5, 7.55,  7.6, 7.65},
    ymin=73.413,
    ymax=73.479,
    xlabel={Latitude $({}^\circ)$},
    label style={font=\footnotesize},
    ytick={73.42, 73.44, 73.46},
    axis background/.style={fill=white},
    xmajorgrids,
    ymajorgrids,
    grid style={opacity=0.5},
    tick label style ={font=\tiny},
    ylabel={Longitude $({}^\circ)$},
    y label style={yshift=-.1cm},
    title style={
        anchor=north,
        at={(0.5, -0.35)},
        font=\footnotesize,
    },
]
\nextgroupplot[
    title={$20\times 20$ \glspl{bf}},
]
\addplot [forget plot] graphics [xmin=7.438, xmax=7.678, ymin=73.413, ymax=73.479] {\currfiledir PureMap20times20BasisFunctions.png};

\nextgroupplot[
    title={$40\times 40$ \glspl{bf}},
    colormap={custom}{
  rgb255(0pt)=(0, 0, 131);
  rgb255(1pt)=(0, 0, 135);
  rgb255(2pt)=(0, 0, 139);
  rgb255(3pt)=(0, 0, 143);
  rgb255(4pt)=(0, 0, 147);
  rgb255(5pt)=(0, 0, 151);
  rgb255(6pt)=(0, 0, 155);
  rgb255(7pt)=(0, 0, 159);
  rgb255(8pt)=(0, 0, 163);
  rgb255(9pt)=(0, 0, 167);
  rgb255(10pt)=(0, 0, 171);
  rgb255(11pt)=(0, 0, 175);
  rgb255(12pt)=(0, 0, 179);
  rgb255(13pt)=(0, 0, 183);
  rgb255(14pt)=(0, 0, 187);
  rgb255(15pt)=(0, 0, 191);
  rgb255(16pt)=(0, 0, 195);
  rgb255(17pt)=(0, 0, 199);
  rgb255(18pt)=(0, 0, 203);
  rgb255(19pt)=(0, 0, 207);
  rgb255(20pt)=(0, 0, 211);
  rgb255(21pt)=(0, 0, 215);
  rgb255(22pt)=(0, 0, 219);
  rgb255(23pt)=(0, 0, 223);
  rgb255(24pt)=(0, 0, 227);
  rgb255(25pt)=(0, 0, 231);
  rgb255(26pt)=(0, 0, 235);
  rgb255(27pt)=(0, 0, 239);
  rgb255(28pt)=(0, 0, 243);
  rgb255(29pt)=(0, 0, 247);
  rgb255(30pt)=(0, 0, 251);
  rgb255(31pt)=(0, 0, 255);
  rgb255(32pt)=(0, 4, 255);
  rgb255(33pt)=(0, 8, 255);
  rgb255(34pt)=(0, 12, 255);
  rgb255(35pt)=(0, 16, 255);
  rgb255(36pt)=(0, 20, 255);
  rgb255(37pt)=(0, 24, 255);
  rgb255(38pt)=(0, 28, 255);
  rgb255(39pt)=(0, 32, 255);
  rgb255(40pt)=(0, 36, 255);
  rgb255(41pt)=(0, 40, 255);
  rgb255(42pt)=(0, 44, 255);
  rgb255(43pt)=(0, 48, 255);
  rgb255(44pt)=(0, 52, 255);
  rgb255(45pt)=(0, 56, 255);
  rgb255(46pt)=(0, 60, 255);
  rgb255(47pt)=(0, 64, 255);
  rgb255(48pt)=(0, 68, 255);
  rgb255(49pt)=(0, 72, 255);
  rgb255(50pt)=(0, 76, 255);
  rgb255(51pt)=(0, 80, 255);
  rgb255(52pt)=(0, 84, 255);
  rgb255(53pt)=(0, 88, 255);
  rgb255(54pt)=(0, 92, 255);
  rgb255(55pt)=(0, 96, 255);
  rgb255(56pt)=(0, 100, 255);
  rgb255(57pt)=(0, 104, 255);
  rgb255(58pt)=(0, 108, 255);
  rgb255(59pt)=(0, 112, 255);
  rgb255(60pt)=(0, 116, 255);
  rgb255(61pt)=(0, 120, 255);
  rgb255(62pt)=(0, 124, 255);
  rgb255(63pt)=(0, 128, 255);
  rgb255(64pt)=(0, 131, 255);
  rgb255(65pt)=(0, 135, 255);
  rgb255(66pt)=(0, 139, 255);
  rgb255(67pt)=(0, 143, 255);
  rgb255(68pt)=(0, 147, 255);
  rgb255(69pt)=(0, 151, 255);
  rgb255(70pt)=(0, 155, 255);
  rgb255(71pt)=(0, 159, 255);
  rgb255(72pt)=(0, 163, 255);
  rgb255(73pt)=(0, 167, 255);
  rgb255(74pt)=(0, 171, 255);
  rgb255(75pt)=(0, 175, 255);
  rgb255(76pt)=(0, 179, 255);
  rgb255(77pt)=(0, 183, 255);
  rgb255(78pt)=(0, 187, 255);
  rgb255(79pt)=(0, 191, 255);
  rgb255(80pt)=(0, 195, 255);
  rgb255(81pt)=(0, 199, 255);
  rgb255(82pt)=(0, 203, 255);
  rgb255(83pt)=(0, 207, 255);
  rgb255(84pt)=(0, 211, 255);
  rgb255(85pt)=(0, 215, 255);
  rgb255(86pt)=(0, 219, 255);
  rgb255(87pt)=(0, 223, 255);
  rgb255(88pt)=(0, 227, 255);
  rgb255(89pt)=(0, 231, 255);
  rgb255(90pt)=(0, 235, 255);
  rgb255(91pt)=(0, 239, 255);
  rgb255(92pt)=(0, 243, 255);
  rgb255(93pt)=(0, 247, 255);
  rgb255(94pt)=(0, 251, 255);
  rgb255(95pt)=(0, 255, 255);
  rgb255(96pt)=(4, 255, 251);
  rgb255(97pt)=(8, 255, 247);
  rgb255(98pt)=(12, 255, 243);
  rgb255(99pt)=(16, 255, 239);
  rgb255(100pt)=(20, 255, 235);
  rgb255(101pt)=(24, 255, 231);
  rgb255(102pt)=(28, 255, 227);
  rgb255(103pt)=(32, 255, 223);
  rgb255(104pt)=(36, 255, 219);
  rgb255(105pt)=(40, 255, 215);
  rgb255(106pt)=(44, 255, 211);
  rgb255(107pt)=(48, 255, 207);
  rgb255(108pt)=(52, 255, 203);
  rgb255(109pt)=(56, 255, 199);
  rgb255(110pt)=(60, 255, 195);
  rgb255(111pt)=(64, 255, 191);
  rgb255(112pt)=(68, 255, 187);
  rgb255(113pt)=(72, 255, 183);
  rgb255(114pt)=(76, 255, 179);
  rgb255(115pt)=(80, 255, 175);
  rgb255(116pt)=(84, 255, 171);
  rgb255(117pt)=(88, 255, 167);
  rgb255(118pt)=(92, 255, 163);
  rgb255(119pt)=(96, 255, 159);
  rgb255(120pt)=(100, 255, 155);
  rgb255(121pt)=(104, 255, 151);
  rgb255(122pt)=(108, 255, 147);
  rgb255(123pt)=(112, 255, 143);
  rgb255(124pt)=(116, 255, 139);
  rgb255(125pt)=(120, 255, 135);
  rgb255(126pt)=(124, 255, 131);
  rgb255(127pt)=(128, 255, 128);
  rgb255(128pt)=(131, 255, 124);
  rgb255(129pt)=(135, 255, 120);
  rgb255(130pt)=(139, 255, 116);
  rgb255(131pt)=(143, 255, 112);
  rgb255(132pt)=(147, 255, 108);
  rgb255(133pt)=(151, 255, 104);
  rgb255(134pt)=(155, 255, 100);
  rgb255(135pt)=(159, 255, 96);
  rgb255(136pt)=(163, 255, 92);
  rgb255(137pt)=(167, 255, 88);
  rgb255(138pt)=(171, 255, 84);
  rgb255(139pt)=(175, 255, 80);
  rgb255(140pt)=(179, 255, 76);
  rgb255(141pt)=(183, 255, 72);
  rgb255(142pt)=(187, 255, 68);
  rgb255(143pt)=(191, 255, 64);
  rgb255(144pt)=(195, 255, 60);
  rgb255(145pt)=(199, 255, 56);
  rgb255(146pt)=(203, 255, 52);
  rgb255(147pt)=(207, 255, 48);
  rgb255(148pt)=(211, 255, 44);
  rgb255(149pt)=(215, 255, 40);
  rgb255(150pt)=(219, 255, 36);
  rgb255(151pt)=(223, 255, 32);
  rgb255(152pt)=(227, 255, 28);
  rgb255(153pt)=(231, 255, 24);
  rgb255(154pt)=(235, 255, 20);
  rgb255(155pt)=(239, 255, 16);
  rgb255(156pt)=(243, 255, 12);
  rgb255(157pt)=(247, 255, 8);
  rgb255(158pt)=(251, 255, 4);
  rgb255(159pt)=(255, 255, 0);
  rgb255(160pt)=(255, 251, 0);
  rgb255(161pt)=(255, 247, 0);
  rgb255(162pt)=(255, 243, 0);
  rgb255(163pt)=(255, 239, 0);
  rgb255(164pt)=(255, 235, 0);
  rgb255(165pt)=(255, 231, 0);
  rgb255(166pt)=(255, 227, 0);
  rgb255(167pt)=(255, 223, 0);
  rgb255(168pt)=(255, 219, 0);
  rgb255(169pt)=(255, 215, 0);
  rgb255(170pt)=(255, 211, 0);
  rgb255(171pt)=(255, 207, 0);
  rgb255(172pt)=(255, 203, 0);
  rgb255(173pt)=(255, 199, 0);
  rgb255(174pt)=(255, 195, 0);
  rgb255(175pt)=(255, 191, 0);
  rgb255(176pt)=(255, 187, 0);
  rgb255(177pt)=(255, 183, 0);
  rgb255(178pt)=(255, 179, 0);
  rgb255(179pt)=(255, 175, 0);
  rgb255(180pt)=(255, 171, 0);
  rgb255(181pt)=(255, 167, 0);
  rgb255(182pt)=(255, 163, 0);
  rgb255(183pt)=(255, 159, 0);
  rgb255(184pt)=(255, 155, 0);
  rgb255(185pt)=(255, 151, 0);
  rgb255(186pt)=(255, 147, 0);
  rgb255(187pt)=(255, 143, 0);
  rgb255(188pt)=(255, 139, 0);
  rgb255(189pt)=(255, 135, 0);
  rgb255(190pt)=(255, 131, 0);
  rgb255(191pt)=(255, 128, 0);
  rgb255(192pt)=(255, 124, 0);
  rgb255(193pt)=(255, 120, 0);
  rgb255(194pt)=(255, 116, 0);
  rgb255(195pt)=(255, 112, 0);
  rgb255(196pt)=(255, 108, 0);
  rgb255(197pt)=(255, 104, 0);
  rgb255(198pt)=(255, 100, 0);
  rgb255(199pt)=(255, 96, 0);
  rgb255(200pt)=(255, 92, 0);
  rgb255(201pt)=(255, 88, 0);
  rgb255(202pt)=(255, 84, 0);
  rgb255(203pt)=(255, 80, 0);
  rgb255(204pt)=(255, 76, 0);
  rgb255(205pt)=(255, 72, 0);
  rgb255(206pt)=(255, 68, 0);
  rgb255(207pt)=(255, 64, 0);
  rgb255(208pt)=(255, 60, 0);
  rgb255(209pt)=(255, 56, 0);
  rgb255(210pt)=(255, 52, 0);
  rgb255(211pt)=(255, 48, 0);
  rgb255(212pt)=(255, 44, 0);
  rgb255(213pt)=(255, 40, 0);
  rgb255(214pt)=(255, 36, 0);
  rgb255(215pt)=(255, 32, 0);
  rgb255(216pt)=(255, 28, 0);
  rgb255(217pt)=(255, 24, 0);
  rgb255(218pt)=(255, 20, 0);
  rgb255(219pt)=(255, 16, 0);
  rgb255(220pt)=(255, 12, 0);
  rgb255(221pt)=(255, 8, 0);
  rgb255(222pt)=(255, 4, 0);
  rgb255(223pt)=(255, 0, 0);
  rgb255(224pt)=(251, 0, 0);
  rgb255(225pt)=(247, 0, 0);
  rgb255(226pt)=(243, 0, 0);
  rgb255(227pt)=(239, 0, 0);
  rgb255(228pt)=(235, 0, 0);
  rgb255(229pt)=(231, 0, 0);
  rgb255(230pt)=(227, 0, 0);
  rgb255(231pt)=(223, 0, 0);
  rgb255(232pt)=(219, 0, 0);
  rgb255(233pt)=(215, 0, 0);
  rgb255(234pt)=(211, 0, 0);
  rgb255(235pt)=(207, 0, 0);
  rgb255(236pt)=(203, 0, 0);
  rgb255(237pt)=(199, 0, 0);
  rgb255(238pt)=(195, 0, 0);
  rgb255(239pt)=(191, 0, 0);
  rgb255(240pt)=(187, 0, 0);
  rgb255(241pt)=(183, 0, 0);
  rgb255(242pt)=(179, 0, 0);
  rgb255(243pt)=(175, 0, 0);
  rgb255(244pt)=(171, 0, 0);
  rgb255(245pt)=(167, 0, 0);
  rgb255(246pt)=(163, 0, 0);
  rgb255(247pt)=(159, 0, 0);
  rgb255(248pt)=(155, 0, 0);
  rgb255(249pt)=(151, 0, 0);
  rgb255(250pt)=(147, 0, 0);
  rgb255(251pt)=(143, 0, 0);
  rgb255(252pt)=(139, 0, 0);
  rgb255(253pt)=(135, 0, 0);
  rgb255(254pt)=(131, 0, 0);
  rgb255(255pt)=(128, 0, 0);
}
,
colorbar horizontal,
colorbar style={height=0.2cm, 
width=\pgfkeysvalueof{/pgfplots/parent axis width}, 
major tick length=0.15em, 
xticklabel pos=top,
xtick={0, 2, 4, 6},
extra x ticks={-2},
extra x tick style={font=\tiny, xticklabel style={yshift=-.025cm}},
every x tick/.style={black},
tick label style={font=\tiny},
minor tick length=0.1em,
minor tick num=1,
at={(parent axis.north)},
anchor=south,
yshift=.5cm},
]
\addplot [forget plot] graphics [xmin=7.438, xmax=7.678, ymin=73.413, ymax=73.479] {\currfiledir PureMap40times40BasisFunctions.png};

\nextgroupplot[
    title={$80\times 80$ \glspl{bf}},
]
\addplot [forget plot] graphics [xmin=7.438, xmax=7.678, ymin=73.413, ymax=73.479] {\currfiledir PureMap80times80BasisFunctions.png};

\end{groupplot}
\end{tikzpicture}%
    \caption{Predictive means indicated by color intensity with transparency proportional to the predictive variance. Increasing number of \glspl{bf} from left to right.}
    \label{fig:magneticpredictivemeans}
    \end{subfigure}\hfill
    \begin{subfigure}{.4\textwidth}\scriptsize
        \begin{tikzpicture}
        \pgfplotsset{set layers}
        \begin{semilogyaxis}[
        scale only axis, 
        axis lines=left,
        height=3cm, 
        width=5cm,
        grid=both,
        legend entries={\textsc{hgp}, Ours},
        legend columns=2,
        legend style={
        at={(0.5, 1.1)},
        anchor=south west,
        draw=none
        },
        ylabel=$\leftarrow$ Wall-clock time $(s)$,
        xlabel=$\leftarrow$ \textsc{nlpd},
        ymin=1e1,
        ymax=1e5,
        tick label style={font=\tiny},
        ]
        \newcommand\spread[2][]{
        \addplot [thick, mark=none, color=#1]
            table [%
            col sep=comma,
            x={MSLL}, y={MeanTimes#2}]
            {\currfiledir /../data/UnderwaterTimingNew.csv};
        \addplot [name path=upper, mark=none, color=#1, dashed, forget plot, opacity=.5]
            table [%
            col sep=comma,
            x={MSLL}, y expr = \thisrow{MeanTimes#2} + \thisrow{STDTimes#2}] %
            {\currfiledir /../data/UnderwaterTimingNew.csv};
        \addplot [name path=lower, mark=none, color=#1, dashed, forget plot, opacity=.5]
            table [%
            col sep=comma,
            x={MSLL}, y expr = \thisrow{MeanTimes#2} - \thisrow{STDTimes#2}] %
            {\currfiledir /../data/UnderwaterTimingNew.csv};
        \addplot[color=#1, fill opacity=0.15, forget plot] fill between[of=lower and upper];
        }
        \spread[Paired-B]{HS}
        \spread[Paired-H]{Hankel}
        \end{semilogyaxis}
        \begin{semilogyaxis}[
        scale only axis,
        separate axis lines,
        x axis line style={-stealth},
        axis x line*=top,
        axis y line=none,
        height=3cm, 
        width=5cm,
        xlabel=Number of \glspl{bf} $(M)$,
        xlabel style={
        anchor=south east,
        },
        x dir=reverse,
        ymin=1e1,
        ymax=1e5,
        xmin=900,
        xmax=6400,
        xtick={900, 3500, 6400},
        tick label style={font=\tiny},
        ]
        \end{semilogyaxis}
    \end{tikzpicture}\\[-10pt]
    \caption{Wall-clock time to compute precision matrix by sequentially including each data point over \glsxtrshort{nlpd}.}
    \label{fig:magneticfielddata}
    \end{subfigure}
\vskip -5pt
\caption{
Our proposed computational scheme reduces the computation time for datasets with high-frequency variations, as these require many \glspl{bf} to achieve accurate reconstruction. This underwater magnetic field has lower \gls{nlpd} with a large amount ($6400$) compared to a smaller amount ($400$) of \glspl{bf}. For $6400$ \glspl{bf}, our computational scheme reduced the required time to compute the precision matrix from 2.7 hours to 1.7 minutes.
}
  \label{fig:magneticfieldpredictions}
\end{figure}
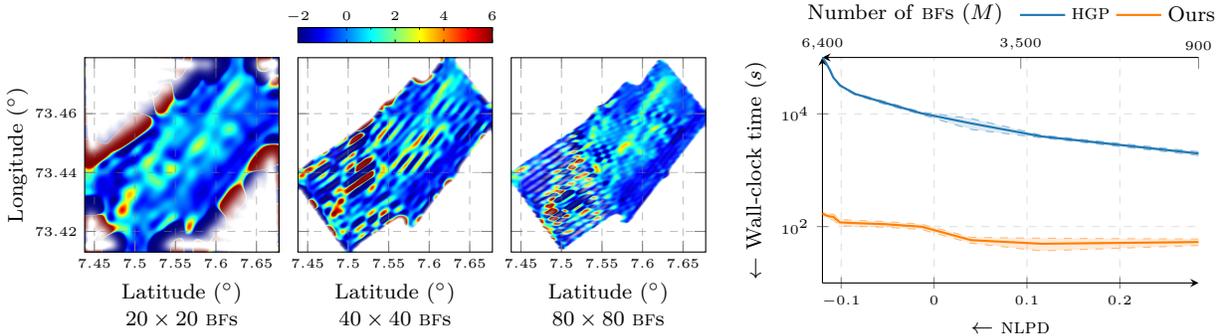

Our first experiment demonstrates the computational scaling of our scheme on a simulated 3D dataset.
Secondly, we consider a magnetic field mapping example with data collected by an underwater vessel, as an application where the high-frequency content of the data requires a large amount of \glspl{bf} to reconstruct the field.
Thirdly, a precipitation dataset is used, mirroring an example in~\citet{solinHilbertSpaceMethods2020}, improving the computational scaling in that particular application even further than the standard \gls{hgp}.
Our freely available \gls{hgp} reference implementation along with hyperparameter optimization is written for GPJax \citep{PinderGPJAX2022}.
All timing experiments are run on an HP Elitebook 840 G5 laptop (Intel i7-8550U CPU, 16GB RAM).
For fair comparison, we naively loop over data points, to avoid any possible low-level optimization skewing the results.

\paragraph{Computational Scaling}
We compare the time necessary for computing the precision matrix for $N=500$ data points for the standard \gls{hgp} as well as for our structure exploiting scheme.
The results are presented in \cref{fig:timing} for an increasing number of \glspl{bf}.
After $M=14000$, the \gls{hgp} is no longer feasible due to memory constraints, whereas our formulation scales well above that, but stop at $M=64000$ for clarity in the illustration.
Clearly, the structure exploitation gives a significantly lower computational cost than the standard \gls{hgp}, even for small quantities of \glspl{bf}.
Further, it drastically reduces the memory requirements, where for $M=14000$, the \gls{hgp} requires roughly $\SI{1.5}{\giga\byte}$ for storing the precision matrix, while our formulation requires roughly $\SI{2.8}{\mega\byte}$ using $64$-bit floats.
This makes it possible for us to scale the number of \glspl{bf} significantly more before running into computational or storage restrictions.
It is noteworthy that even though the implementation is in a high-level language, we still see significant computational savings.

\paragraph{Magnetic Field Mapping}
As \glspl{hgp} are commonly used to model and estimate the magnetic field for, e.g., mapping purposes~\citep{solinMagneticField2015,kokMagneticFieldSLAM2018}, we consider a magnetic field mapping example and demonstrate the ability of our computational scheme to scale \glspl{hgp} to spatially vast datasets.
The data was gathered sequentially in a lawn-mower path in a region approximately $7\times\SI{7}{\kilo\meter}$ large by an underwater vessel outside the coast of Norway ($d=2$ and $N=1.39~\text{million}$).
The data was split into a training set and test set with roughly a $50/50$ split, deterministically split by a grid pattern, to ensure reasonable predictions in the entire data domain, see \cref{app:magneticfield} for more details.
We vary the amount of \glspl{bf} and compare the time required to sequentially include each new data point in the precision matrix as well as the \glsxtrfull{nlpd}.
As the underwater magnetic field covers a large area, a large number of \glspl{bf} are required to accurately represent the field, see \cref{fig:magneticpredictivemeans} where the details of the predicted magnetic field is captured more accurately for an increasing number of \glspl{bf}.
This is also apparent from the decreasing \gls{nlpd} as the number of \glspl{bf} increases, see \cref{fig:magneticfielddata}.
At $6400$ \glspl{bf}, the necessary computation time is several orders of magnitude lower for our approach compared to the standard \gls{hgp}.

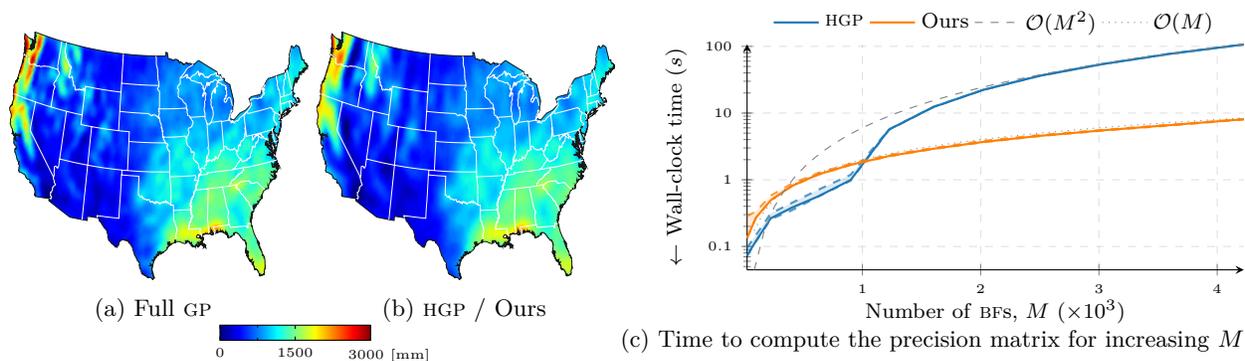
\begin{figure}[t]
    \centering
    \begin{subfigure}[]{.24\textwidth}
        \includestandalone[width=\textwidth]{\currfiledir /../Figures/precipitation/precipitation_gp}
        \caption{Full \gls{gp}
        \label{fig:precipitationfullgp}
        \begin{tikzpicture}[remember picture, overlay, every node/.style={font=\tiny}]
        \pgfplotsset{colormap/jet}
        \begin{axis}[
        hide axis,
        point meta min=0,
        point meta max=30,
        colorbar horizontal,
        colorbar style={height=0.2cm,
        yshift=-.3cm,
        width=2cm,
        major tick length=0.15em, 
        xtick={0, 15},
        xticklabels={$0$, $1500$},
        extra x ticks={30},
        extra x tick labels={$3000$~[mm]},
        extra x tick style={font=\tiny, xticklabel style={xshift=.25cm}},
        every x tick/.style={black},
        tick label style={font=\tiny},
        minor tick length=0.1em,
        minor tick num=1},
        ]
        \end{axis}
    \end{tikzpicture}
        }
    \end{subfigure}
    \begin{subfigure}[]{.24\textwidth}
        \includestandalone[width=\textwidth]{\currfiledir /../Figures/precipitation/precipitation_shgp}
        \caption{\gls{hgp} / Ours}
        \label{fig:precipitationhgp}
    \end{subfigure}
    \begin{subfigure}[]{.5\textwidth}
    \raggedleft
        \begin{tikzpicture}
    \newcommand\spread[2][]{
    \addplot [thick, mark=none, color=#1]
        table [%
        col sep=comma,
        x={m}, y={t_median}]
        {\currfiledir /../data/#2.csv};
    \addplot [thick, name path=upper, mark=none, color=#1, dashed, forget plot, opacity=.75]
        table [%
        col sep=comma,
        x={m}, y={t_max}] %
        {\currfiledir /../data/#2.csv};
    \addplot [thick, name path=lower, mark=none, color=#1, dashed, forget plot, opacity=.75]
        table [%
        col sep=comma,
        x={m}, y={t_min}] %
        {\currfiledir /../data/#2.csv};
    \addplot[color=#1, fill opacity=0.15, forget plot] fill between[of=lower and upper];
    }
        \begin{semilogyaxis}[
        scale only axis, 
        axis lines=left,
        height=3cm, 
        width=.8\textwidth,
        grid=major,
        legend entries={\textsc{hgp}, Ours, $\mathcal{O}(M^2)$, $\mathcal{O}(M)$},
        legend columns=4,
        legend style={
        at={(0.5, 1.0)},
        anchor=south,
        draw=none
        },
        ylabel=$\leftarrow$ Wall-clock time $(s)$,
        log ticks with fixed point,
        scaled x ticks=base 10:-3,
        xlabel={Number of \textsc{bf}s, $M~({\times}10^3)$},
        x label style={yshift=.15cm},
        xtick scale label code/.code={},
        y tick label style={font=\tiny},
        x tick label style={font=\tiny},
        ]
        \spread[Paired-B]{HGPprecipitation_timing}
        \spread[Paired-H]{SHGPprecipitation_timing}
        \addplot [samples=50, color=gray, domain=0:4225, dashed] {6e-6 * x^2};
        \addplot [samples=50, color=gray, domain=0:4225, dotted] {2e-3 * x};
        \end{semilogyaxis}
    \end{tikzpicture}\\[-8pt]
    \caption{Time to compute the precision matrix for increasing $M$}
    \label{fig:precipitationtiming}
    \end{subfigure}\vskip -5pt
    \caption{These experiments recover the results from~\citet{solinHilbertSpaceMethods2020} exactly for predicting yearly precipitation levels across the US, and measure the wall-clock time needed by our proposed computational scheme. The \gls{hgp} efficiently approximates the full \gls{gp} solution using $m_1=m_2=45$, totaling $M=2025$ \glspl{bf}.}
    \label{fig:precipitationpredictions}
\end{figure}

\paragraph{U.S.\ Precipitation Data}
We consider a standard precipitation data set containing US annual precipitation summaries for year 1995 ($d=2$ and $N=5776$) \citep{vanhataloPrecipitation2008}.
We exactly mimic the setup in \citet{solinHilbertSpaceMethods2020} and primarily focus our evaluation on the calculation of the precision matrix.
The time for computing the precision matrix is visualized in \cref{fig:precipitationtiming}, where our approach clearly outperforms the standard \gls{hgp}.
The predictions on a dense grid over the continental US can be found in \cref{fig:precipitationfullgp,fig:precipitationhgp}, where the \gls{hgp} manages to capture both the large-scale as well as the small-scale variations well.

\section{Related Work}\label{sec:relatedwork}

The computational scheme that we detail here can be used to speed up a range of approximate \gls{gp} and kernel methods with \glspl{bf} that satisfy the Hankel--Toeplitz structure we use~\citep[see e.g.][]{Tompkins_Ramos_2018, solinHilbertSpaceMethods2020, hensmanVariational2017}. 
It is also applicable to~\citet{lazaro-gredilla_sparse_spectrum} in the special case where the considered frequencies of the \glspl{bf} are equidistant, as well as~\citet{pmlr-v119-dutordoir20a} when the input space is one-dimensional. 
However, there is also a wide range of \gls{gp} approximations that do not have the Hankel--Toeplitz structure required for~\cref{thm:2md} or~\cref{thm:3md} to apply.
While the structure we exploit in~\cref{thm:3md} is apparent in the quadrature Fourier feature approach of~\citet{Mutn2018EfficientHDQuadratureFourier} when the frequencies are on a structured Cartesian grid, other quadrature-like methods are not possible to speed-up in similar ways.
The most well-known method is the random Fourier feature approach~\citep{rahimi_rff_2007}, where speed-up is not possible as the frequencies are sampled at random.
Similarly, the Gaussian quadrature approaches of~\citet{dao_qff_2017,shustin_gausslegendre_2022} and the random quadrature approach of~\citet{munkhoeva_quadrature_2018} do not constrain frequencies to a regular grid and therefore do not have the Hankel--Toeplitz structure required by~\cref{thm:3md}.

As was pointed out by \citet{quinonero-candelaUnifyingViewSparse2005}, inducing point approaches can also be viewed as \gls{bf} approximations. 
The inducing variable approaches essentially summarize the data in a set of inducing variables~\citep{snelsonSparseGaussianProcesses2005,seegerFastForwardSelection2003,csatoSparseOnLineGaussian2002}, which most commonly represent function values at a certain set of inputs, even though other choices are possible~\citep{lazaro-gredillaInterdomainGaussianProcesses2009}.  
Another closely related approach which also uses inducing points, with similar computational complexity of $\mathcal{O}(NM^2)$, is the variational \gls{gp} pioneered by~\citet{titsiasVariationalLearningInducing}. 
In light of the \gls{bf} viewpoint, all the aforementioned approaches therefore also involve computing a precision matrix, but even for ordered inducing points on a grid, the particular Hankel--Toeplitz structure we exploit does not exist in general. \looseness-1

A range of related work stacks additional approximations on top of the sparse basis function/inducing point approximations~\citep{yadav_faster_2021, izmailovScalableGaussianProcesses2018, hensmanGaussianProcessesBig2013}. The purpose of these additional approximations is typically to either reduce the computational complexity of computing the precision matrix for example by using \glspl{bf} with compact support, or by implementing an approximate algorithm for inverting the precision matrix. 
A well-known structure exploiting method is \gls{ski} \citep{wilsonKernelInterpolationScalable,yadav_faster_2021,izmailovScalableGaussianProcesses2018}, which approximates the precision matrix through cubic interpolation between inducing points on a regular grid. 
An alternative approach that improves the computational complexity of the variational \gls{gp}, is the \gls{svgp} \citep{hensmanGaussianProcessesBig2013}. 
The \gls{svgp} is the {\it de~facto} standard approach for large--scale \glspl{gp}, due to the possibility of utilizing mini-batching for training, greatly speeding up hyperparameter (and variational parameter) learning~\citep{hensmanGaussianProcessesBig2013,hensmanScalableVariational2015}, with implementation available in, e.g., GPyTorch and GPflow~\citep{gardnerGPyTorchBlackboxMatrixMatrix2018,MatthewsGPFLOW2017}. 

While this paper focuses on \gls{bf} approximations to \glspl{gp}, there is a lot of work on approximating exact \gls{gp} regression through means of, e.g., \gls{cg} descent~\citep{gibbsmackayConjugateGradient1996,artemevTighterBoundsLog2021,gardnerGPyTorchBlackboxMatrixMatrix2018}.
These methods view \gls{gp} regression as the solution to a linear system of equations and seek to solve this approximately.
This typically reduces the cost of exact \gls{gp} regression from $\mathcal{O}(N^3)$ to $\mathcal{O}(IN^2)$, where $I$ is the number of \gls{cg} iterations.
\citet{daviesEffectiveCG2015} uses \gls{cg} as the driving scalability mechanism, but further reduces the computational complexity to $\mathcal{O}(NMI)$ through use of \emph{M-efficient kernels}, which essentially constitute the approaches from \citet{quinonero-candelaUnifyingViewSparse2005} as well as compact kernels \citep[see, e.g.,][]{kullbergOnlineJointState2021,BuhmannRBF2003,WuCompactBF1995}.
\gls{cg} has also been combined with \gls{ski} in the \textsc{kiss-gp} framework of \citet{wilsonKernelInterpolationScalable} to reduce the computational complexity of (mean) inference to $\mathcal{O}(N+M\log M)$. 
\citet{pleissConstantTimePredictiveDistributions2018a} extended \textsc{kiss-gp} with the Lanczos algorithm to reduce the complexity of computing the predictive variance to $\mathcal{O}(k)$ after pre-computation, where $k$ is the number of Lanczos iterations.
Recently, \gls{sgd} was introduced as an alternative to \gls{cg} \citep{linSamplingGPSGD2023,linStochasticGradientDescent2023}, potentially with better performance for ill-conditioned datasets.
Technically, any of these ideas are straightforward to include in the \gls{hgp} as well and could potentially be used to reduce the $\mathcal{O}(M^3)$ complexity of inference and hyperparameter learning.

Other previous work has discovered that functions of difference matrices (a special case of Toeplitz matrices) are also sums of Hankel and Toeplitz matrices in the one-dimensional case, and Kronecker products of these in the two-dimensional case~\citep{heat_eq_paper2014}.
Since our matrices are also Kronecker products of Hankel and Toeplitz matrices, we end up exploiting similar structures as~\citet{heat_eq_paper2014} arising in a different situation.\looseness-1

\section{Conclusion}\label{sec:discussion}

Our contribution details a computational approach for exploiting Hankel and Toeplitz structures that appear in multiple \gls{bf} approximation schemes to kernels for \glspl{gp}.
These structures allow us to reduce the computational complexity of computing the corresponding precision matrix from $\mathcal{O}(NM^2)$ to $\mathcal{O}(NM)$ \emph{without} further approximations.
Further, our approach reduces the storage requirement for containing all necessary information about the posterior to make predictions from $\mathcal{O}(M^2)$ to $\mathcal{O}(M)$.
The Hankel and Toeplitz structures appear because of the particular \glspl{bf} that are used to approximate the kernel, \emph{not} the kernel itself.
The reduced computational and storage requirements are particularly beneficial in the \gls{hgp} where more \glspl{bf} allow us to capture higher frequencies of the kernel spectrum, otherwise unattainable without significant computational resources.
We foresee that our contribution will allow \glspl{hgp} to tackle larger problems without the need for extensive specialized hardware, opening up approximate \gls{gp} learning and inference for a wider audience. 
Future work could investigate if the results can be generalized to wider ranges of \glspl{bf}, which is easily verified through \cref{thm:2md,thm:3md}. Another potential is to investigate ways of approximately decomposing an already known precision matrix into Hankel--Toeplitz matrices which does not admit an analytical such decomposition.
This would yield the same computational benefits but with some potential loss of accuracy.

A reference implementation built on top of GPJax is available at \href{https://github.com/AOKullberg/hgp-hankel-structure}{https://github.com/AOKullberg/hgp-hankel-structure}.

\section*{Impact Statement}
This work develops numerical methods for the wide field of machine learning, where the goal is to make existing methods more compute-efficient and open new avenues for developing future methods. There are many potential uses and therefore societal consequences of such methods, none of which we see the need to specifically highlight here.

\section*{Acknowledgements}
We would like to express our gratitude to the supervisors (Gustaf Hendeby, Isaac Skog, Kim Batselier, Manon Kok and Rudy Helmons) of the three PhD candidates (Anton Kullberg, Frederiek Wesel and Frida Viset) involved in this project. They secured the funding that made this research possible. Their support in providing the necessary resources and their encouragement for our development as independent researchers have been invaluable. Their contributions have thus indirectly shaped this work, and we are grateful for their continued guidance and support.
 We would like to thank the anonymous reviewers for their numerous suggestions which have greatly improved the quality of this paper. 
 Frederiek Wesel, and thereby this work, is supported by the Delft University of Technology AI Labs program. 
 Arno Solin acknowledges funding from the Research Council of Finland (grant id 339730).
 The underwater magnetic field data used were collected by MARMINE/NTNU research cruise funded by the Research Council of Norway (Norges Forskningsråd, NFR) Project No.\ 247626/O30 and associated industrial partners. Ocean Floor Geophysics provided the magnetometer that was used for magnetic data acquisition and pre‐processed the magnetic data. The authors declare no competing interests.

\bibliography{main}

\begin{thebibliography}{51}
\providecommand{\natexlab}[1]{#1}
\providecommand{\url}[1]{\texttt{#1}}
\expandafter\ifx\csname urlstyle\endcsname\relax
  \providecommand{\doi}[1]{doi: #1}\else
  \providecommand{\doi}{doi: \begingroup \urlstyle{rm}\Url}\fi

\bibitem[Abril-Pla et~al.(2023)Abril-Pla, Andreani, Carroll, Dong, Fonnesbeck,
  Kochurov, Kumar, Lao, Luhmann, Martin, et~al.]{pymc}
Oriol Abril-Pla, Virgile Andreani, Colin Carroll, Larry Dong, Christopher~J
  Fonnesbeck, Maxim Kochurov, Ravin Kumar, Junpeng Lao, Christian~C Luhmann,
  Osvaldo~A Martin, et~al.
\newblock {PyMC}: a modern, and comprehensive probabilistic programming
  framework in {P}ython.
\newblock \emph{PeerJ Computer Science}, 9:\penalty0 e1516, 2023.

\bibitem[Artemev et~al.(2021)Artemev, Burt, and van~der
  Wilk]{artemevTighterBoundsLog2021}
Artem Artemev, David~R. Burt, and Mark van~der Wilk.
\newblock Tighter bounds on the log marginal likelihood of {G}aussian process
  regression using conjugate gradients.
\newblock In \emph{Proceedings of the 38th International Conference on Machine
  Learning}, volume 139 of \emph{Proceedings of Machine Learning Research},
  pp.\  362--372. {PMLR}, 2021.

\bibitem[Berntorp(2021)]{berntorpOnlineBayesianInference2021a}
Karl Berntorp.
\newblock Online {B}ayesian inference and learning of
  {G}aussian{\textendash}process state{\textendash}space models.
\newblock \emph{Automatica}, 129:\penalty0 109613, July 2021.

\bibitem[Buhmann(2003)]{BuhmannRBF2003}
Martin~D. Buhmann.
\newblock \emph{{Radial Basis Functions: Theory and Implementations}}.
\newblock Cambridge University Press, 2003.

\bibitem[Chen et~al.(2018)Chen, Batselier, Suykens, and
  Wong]{chen_parallelized_2018}
Zhongming Chen, Kim Batselier, Johan A.~K. Suykens, and Ngai Wong.
\newblock Parallelized tensor train learning of polynomial classifiers.
\newblock \emph{IEEE Transactions on Neural Networks and Learning Systems},
  29\penalty0 (10):\penalty0 4621--4632, 2018.

\bibitem[Csat{\'o} \& Opper(2002)Csat{\'o} and
  Opper]{csatoSparseOnLineGaussian2002}
Lehel Csat{\'o} and Manfred Opper.
\newblock Sparse on-line {G}aussian processes.
\newblock \emph{Neural Computation}, 14\penalty0 (3):\penalty0 641--668, March
  2002.

\bibitem[Dao et~al.(2017)Dao, De~Sa, and R\'{e}]{dao_qff_2017}
Tri Dao, Christopher~M De~Sa, and Christopher R\'{e}.
\newblock {G}aussian quadrature for kernel features.
\newblock In \emph{Advances in Neural Information Processing Systems},
  volume~30. Curran Associates, Inc., 2017.

\bibitem[Davies(2015)]{daviesEffectiveCG2015}
Alexander~James Davies.
\newblock \emph{Effective {I}mplementation of {G}aussian {P}rocess {R}egression
  for {M}achine {L}earning}.
\newblock PhD thesis, University of Cambridge, 2015.

\bibitem[de~G.~Matthews et~al.(2017)de~G.~Matthews, van~der Wilk, Nickson,
  Fujii, Boukouvalas, Le{\'o}n-Villagr{\'a}, Ghahramani, and
  Hensman]{MatthewsGPFLOW2017}
Alexander~G. de~G.~Matthews, Mark van~der Wilk, Tom Nickson, Keisuke Fujii,
  Alexis Boukouvalas, Pablo Le{\'o}n-Villagr{\'a}, Zoubin Ghahramani, and James
  Hensman.
\newblock {GPflow: A Gaussian process library using TensorFlow}.
\newblock \emph{Journal of Machine Learning Research}, 18\penalty0
  (40):\penalty0 1--6, 2017.

\bibitem[Dutordoir et~al.(2020)Dutordoir, Durrande, and
  Hensman]{pmlr-v119-dutordoir20a}
Vincent Dutordoir, Nicolas Durrande, and James Hensman.
\newblock Sparse {G}aussian processes with spherical harmonic features.
\newblock In \emph{Proceedings of the 37th International Conference on Machine
  Learning}, volume 119 of \emph{Proceedings of Machine Learning Research},
  pp.\  2793--2802. PMLR, 2020.

\bibitem[Gardner et~al.(2018)Gardner, Pleiss, Weinberger, Bindel, and
  Wilson]{gardnerGPyTorchBlackboxMatrixMatrix2018}
Jacob Gardner, Geoff Pleiss, Kilian~Q Weinberger, David Bindel, and Andrew~G
  Wilson.
\newblock {{GPyTorch}}: {{Blackbox matrix-matrix Gaussian process inference}}
  with {{GPU acceleration}}.
\newblock In \emph{Advances in Neural Information Processing Systems},
  volume~31. {Curran Associates, Inc.}, 2018.

\bibitem[Gibbs \& MacKay(1996)Gibbs and
  MacKay]{gibbsmackayConjugateGradient1996}
Mark~N. Gibbs and David J.~C. MacKay.
\newblock Efficient implementation of {G}aussian processes for interpolation.
\newblock Technical Report, University of Cambridge, 1996.

\bibitem[Greengard et~al.(2023)Greengard, Rachh, and
  Barnett]{greengard2023equispaced}
Philip Greengard, Manas Rachh, and Alex Barnett.
\newblock Equispaced fourier representations for efficient gaussian process
  regression from a billion data points.
\newblock arXiv, 2023.

\bibitem[Hensman et~al.(2013)Hensman, Fusi, and
  Lawrence]{hensmanGaussianProcessesBig2013}
James Hensman, Nicol\`{o} Fusi, and Neil~D. Lawrence.
\newblock {G}aussian processes for big data.
\newblock In \emph{Proceedings of the 29th Conference on Uncertainty in
  Artificial Intelligence}, UAI, pp.\  282–290. AUAI Press, 2013.

\bibitem[Hensman et~al.(2015)Hensman, Matthews, and
  Ghahramani]{hensmanScalableVariational2015}
James Hensman, Alexander Matthews, and Zoubin Ghahramani.
\newblock Scalable variational {G}aussian process classification.
\newblock In \emph{Proceedings of the 18th International Conference on
  Artificial Intelligence and Statistics}, volume~38 of \emph{Proceedings of
  Machine Learning Research}, pp.\  351--360. PMLR, 2015.

\bibitem[Hensman et~al.(2017)Hensman, Durrande, and
  Solin]{hensmanVariational2017}
James Hensman, Nicolas Durrande, and Arno Solin.
\newblock Variational {F}ourier features for {G}aussian processes.
\newblock \emph{The Journal of Machine Learning Research}, 18\penalty0
  (1):\penalty0 5537--5588, January 2017.

\bibitem[Izmailov et~al.(2018)Izmailov, Novikov, and
  Kropotov]{izmailovScalableGaussianProcesses2018}
Pavel Izmailov, Alexander Novikov, and Dmitry Kropotov.
\newblock Scalable {G}aussian processes with billions of inducing inputs via
  tensor train decomposition.
\newblock In \emph{Proceedings of the 21st International Conference on
  Artificial Intelligence and Statistics}, volume~84 of \emph{Proceedings of
  Machine Learning Research}, pp.\  726--735. PMLR, 2018.

\bibitem[Jang et~al.(2020)Jang, Yoo, Son, Kim, and Kim]{jang_multi-robot_2020}
Dohyun Jang, Jaehyun Yoo, Clark~Youngdong Son, Dabin Kim, and H.~Jin Kim.
\newblock Multi{\textendash}robot active sensing and environmental model
  learning with distributed {G}aussian process.
\newblock \emph{IEEE Robotics and Automation Letters}, 5\penalty0 (4):\penalty0
  5905--5912, 2020.

\bibitem[Kingma \& Adam(2015)Kingma and Adam]{kingma2015Adam}
Diederik~P. Kingma and Jimmy~Ba. Adam.
\newblock Adam: {A} method for stochastic optimization.
\newblock In \emph{Proceedings of the 3rd International Conference on Learning
  Representations ({ICLR})}, 2015.

\bibitem[Kok \& Solin(2018)Kok and Solin]{kokMagneticFieldSLAM2018}
Manon Kok and Arno Solin.
\newblock Scalable magnetic field {SLAM} in {3D} using {G}aussian process maps.
\newblock In \emph{Proceedings of the 21st International Conference on
  Information Fusion (FUSION)}, pp.\  1353--1360, 2018.

\bibitem[Kullberg et~al.(2021)Kullberg, Skog, and
  Hendeby]{kullbergOnlineJointState2021}
Anton Kullberg, Isaac Skog, and Gustaf Hendeby.
\newblock Online joint state inference and learning of partially unknown
  state{\textendash}space models.
\newblock \emph{IEEE Transactions on Signal Processing}, 69:\penalty0
  4149--4161, 2021.

\bibitem[{L{\'a}zaro-Gredilla} \& {Figueiras-Vidal}(2009){L{\'a}zaro-Gredilla}
  and {Figueiras-Vidal}]{lazaro-gredillaInterdomainGaussianProcesses2009}
Miguel {L{\'a}zaro-Gredilla} and An{\'i}bal {Figueiras-Vidal}.
\newblock Inter-domain {G}aussian processes for sparse inference using inducing
  features.
\newblock In \emph{Advances in Neural Information Processing Systems.},
  volume~22, 2009.

\bibitem[L{{\'a}}zaro-Gredilla et~al.(2010)L{{\'a}}zaro-Gredilla,
  Qui{{\~n}}onero-Candela, Rasmussen, and
  Figueiras-Vidal]{lazaro-gredilla_sparse_spectrum}
Miguel L{{\'a}}zaro-Gredilla, Joaquin Qui{{\~n}}onero-Candela, Carl~Edward
  Rasmussen, and An{\'i}bal~R. Figueiras-Vidal.
\newblock Sparse spectrum {G}aussian process regression.
\newblock \emph{Journal of Machine Learning Research}, 11\penalty0
  (63):\penalty0 1865--1881, 2010.

\bibitem[Lin et~al.(2023{\natexlab{a}})Lin, Antor{\'a}n, Padhy, Janz,
  Hern{\'a}ndez-Lobato, and Terenin]{linSamplingGPSGD2023}
Jihao~Andreas Lin, Javier Antor{\'a}n, Shreyas Padhy, David Janz,
  Jos{\'e}~Miguel Hern{\'a}ndez-Lobato, and Alexander Terenin.
\newblock {Sampling from Gaussian process posteriors using stochastic gradient
  descent}.
\newblock In \emph{Advances in Neural Information Processing Systems},
  volume~37, 2023{\natexlab{a}}.

\bibitem[Lin et~al.(2023{\natexlab{b}})Lin, Padhy, Antor{\'a}n, Tripp, Terenin,
  Szepesv{\'a}ri, {Hern{\'a}ndez-Lobato}, and
  Janz]{linStochasticGradientDescent2023}
Jihao~Andreas Lin, Shreyas Padhy, Javier Antor{\'a}n, Austin Tripp, Alexander
  Terenin, Csaba Szepesv{\'a}ri, Jos{\'e}~Miguel {Hern{\'a}ndez-Lobato}, and
  David Janz.
\newblock Stochastic gradient descent for {G}aussian processes done right.
\newblock arXiv:2310.20581, 2023{\natexlab{b}}.

\bibitem[Lindgren et~al.(2022)Lindgren, Bolin, and Rue]{lindgrenSpde2022}
Finn Lindgren, David Bolin, and H{\aa}vard Rue.
\newblock The {{SPDE}} approach for {{Gaussian}} and non-{{Gaussian}} fields:
  10 {Y}ears and still running.
\newblock \emph{Spatial Statistics}, 50:\penalty0 100599, August 2022.

\bibitem[Munkhoeva et~al.(2018)Munkhoeva, Kapushev, Burnaev, and
  Osele\~dets]{munkhoeva_quadrature_2018}
Marina Munkhoeva, Yermek Kapushev, Evgeny Burnaev, and Ivan Osele\~dets.
\newblock Quadrature-based features for kernel approximation.
\newblock In \emph{Advances in Neural Information Processing Systems},
  volume~31. Curran Associates, Inc., 2018.

\bibitem[Mutn{\'y} \& Krause(2018)Mutn{\'y} and
  Krause]{Mutn2018EfficientHDQuadratureFourier}
Mojm{\'i}r Mutn{\'y} and Andreas Krause.
\newblock Efficient high dimensional {{Bayesian}} optimization with additivity
  and quadrature {F}ourier features.
\newblock In \emph{Advances in Neural Information Processing Systems},
  volume~32, pp.\  9019--9030, 2018.

\bibitem[Novikov et~al.(2018)Novikov, Trofimov, and
  Oseledets]{novikov_exponential_2018}
A.~Novikov, M.~Trofimov, and I.~Oseledets.
\newblock Exponential machines.
\newblock \emph{Bulletin of the Polish Academy of Sciences Technical Sciences},
  66\penalty0 (6 (Special Section on Deep Learning: Theory and
  Practice)):\penalty0 789--797, 2018.

\bibitem[Pillonetto et~al.(2019)Pillonetto, Schenato, and
  Varagnolo]{pillonetto_distributed_2018}
Gianluigi Pillonetto, Luca Schenato, and Damiano Varagnolo.
\newblock Distributed multi{\textendash}agent {G}aussian regression via
  finite{\textendash}dimensional approximations.
\newblock \emph{IEEE Transactions on Pattern Analysis and Machine
  Intelligence}, 41\penalty0 (9):\penalty0 2098--2111, 2019.

\bibitem[Pinder \& Dodd(2022)Pinder and Dodd]{PinderGPJAX2022}
Thomas Pinder and Daniel Dodd.
\newblock {GPJax}: {A} {G}aussian process framework in {JAX}.
\newblock \emph{Journal of Open Source Software}, 7\penalty0 (75):\penalty0
  4455, 2022.

\bibitem[Pleiss et~al.(2018)Pleiss, Gardner, Weinberger, and
  Wilson]{pleissConstantTimePredictiveDistributions2018a}
Geoff Pleiss, Jacob Gardner, Kilian Weinberger, and Andrew~Gordon Wilson.
\newblock Constant-time predictive distributions for {G}aussian processes.
\newblock In \emph{Proceedings of the 35th International Conference on Machine
  Learning}, volume~80 of \emph{Proceedings of Machine Learning Research}, pp.\
   4114--4123. PMLR, 2018.

\bibitem[Qui{{\~n}}onero-Candela \& Rasmussen(2005)Qui{{\~n}}onero-Candela and
  Rasmussen]{quinonero-candelaUnifyingViewSparse2005}
Joaquin Qui{{\~n}}onero-Candela and Carl~Edward Rasmussen.
\newblock A unifying view of sparse approximate {G}aussian process regression.
\newblock \emph{Journal of Machine Learning Research}, 6\penalty0
  (65):\penalty0 1939--1959, 2005.

\bibitem[Rahimi \& Recht(2007)Rahimi and Recht]{rahimi_rff_2007}
Ali Rahimi and Benjamin Recht.
\newblock Random features for large-scale kernel machines.
\newblock In \emph{Advances in Neural Information Processing Systems},
  volume~20. Curran Associates, Inc., 2007.

\bibitem[Rasmussen \& Williams(2006)Rasmussen and
  Williams]{rasmussen2005gaussian}
C.E. Rasmussen and C.K.I. Williams.
\newblock \emph{Gaussian Processes for Machine Learning}.
\newblock Adaptive Computation and Machine Learning series. MIT Press, 2006.

\bibitem[{Riutort-Mayol} et~al.(2022){Riutort-Mayol}, B{\"u}rkner, Andersen,
  Solin, and Vehtari]{riutortPractical2022}
Gabriel {Riutort-Mayol}, Paul-Christian B{\"u}rkner, Michael~R. Andersen, Arno
  Solin, and Aki Vehtari.
\newblock Practical {{Hilbert}} space approximate {{Bayesian Gaussian}}
  processes for probabilistic programming.
\newblock \emph{Statistics and Computing}, 33\penalty0 (1):\penalty0 17,
  December 2022.

\bibitem[Seeger et~al.(2003)Seeger, Williams, and
  Lawrence]{seegerFastForwardSelection2003}
Matthias~W. Seeger, Christopher K.~I. Williams, and Neil~D. Lawrence.
\newblock Fast forward selection to speed up sparse {G}aussian process
  regression.
\newblock In \emph{International Workshop on Artificial Intelligence and
  Statistics}, volume~9 of \emph{Proceedings of Machine Learning Research},
  pp.\  254--261. {PMLR}, 2003.

\bibitem[Shustin \& Avron(2022)Shustin and Avron]{shustin_gausslegendre_2022}
Paz~Fink Shustin and Haim Avron.
\newblock {G}auss-{L}egendre features for {G}aussian process regression.
\newblock \emph{Journal of Machine Learning Research}, 23\penalty0
  (92):\penalty0 1--47, 2022.

\bibitem[Snelson \& Ghahramani(2005)Snelson and
  Ghahramani]{snelsonSparseGaussianProcesses2005}
Edward Snelson and Zoubin Ghahramani.
\newblock Sparse {G}aussian processes using pseudo-inputs.
\newblock In \emph{Advances in Neural Information Processing Systems},
  volume~18. MIT Press, 2005.

\bibitem[Solin \& S{\"a}rkk{\"a}(2020)Solin and
  S{\"a}rkk{\"a}]{solinHilbertSpaceMethods2020}
Arno Solin and Simo S{\"a}rkk{\"a}.
\newblock Hilbert space methods for reduced-rank {G}aussian process regression.
\newblock \emph{Statistics and Computing}, 30\penalty0 (2):\penalty0 419--446,
  March 2020.

\bibitem[Solin et~al.(2015)Solin, Kok, Wahlström, Schön, and
  Särkkä]{solinMagneticField2015}
Arno Solin, Manon Kok, Niklas Wahlström, Thomas~B Schön, and Simo Särkkä.
\newblock Modeling and interpolation of the ambient magnetic field by
  {G}aussian processes.
\newblock \emph{IEEE Transactions on Robotics}, 34\penalty0 (4):\penalty0
  1112--1127, September 2015.

\bibitem[Strang \& MacNamara(2014)Strang and MacNamara]{heat_eq_paper2014}
Gilbert Strang and Shev MacNamara.
\newblock Functions of difference matrices are {T}oeplitz plus {H}ankel.
\newblock \emph{SIAM Review}, 56\penalty0 (3):\penalty0 525--546, 2014.

\bibitem[Svensson et~al.(2016)Svensson, Solin, Särkkä, and
  Schön]{svenssonComputationallyEfficientBayesian2016}
Andreas Svensson, Arno Solin, Simo Särkkä, and Thomas Schön.
\newblock Computationally efficient {B}ayesian learning of {G}aussian process
  state space models.
\newblock In \emph{Proceedings of the 19th International Conference on
  Artificial Intelligence and Statistics}, volume~51, pp.\  213--221. PMLR,
  2016.

\bibitem[Titsias(2009)]{titsiasVariationalLearningInducing}
Michalis Titsias.
\newblock Variational learning of inducing variables in sparse {G}aussian
  processes.
\newblock In \emph{Proceedings of the 12th International Conference on
  Artificial Intelligence and Statistics}, volume~5 of \emph{Proceedings of
  Machine Learning Research}, pp.\  567--574. PMLR, 2009.

\bibitem[Tompkins \& Ramos(2018)Tompkins and Ramos]{Tompkins_Ramos_2018}
Anthony Tompkins and Fabio Ramos.
\newblock Fourier feature approximations for periodic kernels in time-series
  modelling.
\newblock \emph{Proceedings of the AAAI Conference on Artificial Intelligence},
  32\penalty0 (1), 2018.

\bibitem[Vanhatalo \& Vehtari(2008)Vanhatalo and
  Vehtari]{vanhataloPrecipitation2008}
Jarno Vanhatalo and Aki Vehtari.
\newblock Modelling local and global phenomena with sparse {G}aussian
  processes.
\newblock In \emph{Proceedings of the 24th Conference on Uncertainty in
  Artificial Intelligence (UAI)}, pp.\  571 – 578. AUAI Press, 2008.

\bibitem[Viset et~al.(2023)Viset, Helmons, and
  Kok]{viset2023DistributedMagSLAM}
Frida Viset, Rudy Helmons, and Manon Kok.
\newblock Distributed multi-agent magnetic field norm slam with gaussian
  processes.
\newblock In \emph{2023 26th International Conference on Information Fusion
  (FUSION)}, pp.\  1--8, 2023.

\bibitem[Wahls et~al.(2014)Wahls, Koivunen, Poor, and
  Verhaegen]{wahls_learning_2014}
Sander Wahls, Visa Koivunen, H.~Vincent Poor, and Michel Verhaegen.
\newblock Learning multidimensional {F}ourier series with tensor trains.
\newblock In \emph{2014 IEEE Global Conference on Signal and Information
  Processing (GlobalSIP)}, pp.\  394--398. IEEE, 2014.

\bibitem[Wilson \& Nickisch(2015)Wilson and
  Nickisch]{wilsonKernelInterpolationScalable}
Andrew Wilson and Hannes Nickisch.
\newblock Kernel interpolation for scalable structured {G}aussian processes
  {(KISS{\textendash}GP)}.
\newblock In \emph{Proceedings of the 32nd International Conference on Machine
  Learning}, volume~37 of \emph{Proceedings of Machine Learning Research}, pp.\
   1775--1784. PMLR, 2015.

\bibitem[Wu(1995)]{WuCompactBF1995}
Zongmin Wu.
\newblock Compactly supported positive definite radial functions.
\newblock \emph{Advances in Computational Mathematics}, 4\penalty0
  (1):\penalty0 283--292, 1995.

\bibitem[Yadav et~al.(2021)Yadav, Sheldon, and Musco]{yadav_faster_2021}
Mohit Yadav, Daniel Sheldon, and Cameron Musco.
\newblock Faster kernel interpolation for {G}aussian processes.
\newblock In \emph{Proceedings of The 24th International Conference on
  Artificial Intelligence and Statistics}, volume 130 of \emph{Proceedings of
  Machine Learning Research}, pp.\  2971--2979. PMLR, 2021.

\end{thebibliography}
\bibliographystyle{tmlr}

\newpage
\appendix
\onecolumn

\section*{Appendix}

\renewcommand{\thetable}{A\arabic{table}}
\renewcommand{\thefigure}{A\arabic{figure}}

The majoriy of the appendix covers proofs of the corollaries following from \cref{thm:2md,thm:3md}.
The rest is dedicated to further details on our empirical experiments as well as visual explanations of the structures that are exploited in the main body of the paper.
The appendix is organized as follows.
\cref{app:polynomial} proves \cref{thm:2md} for polynomial \glsfirstplural{bf}.
\cref{app:complexexponentialproof} proves \cref{thm:2md} for complex exponential \glspl{bf}.
\cref{app:hgprectangular} proves \cref{thm:3md} for Hilbert space \glspl{bf} defined on a rectangular domain.
\cref{app:ordinatyFourierFeatures} proves \cref{thm:3md} for ordinary Fourier features.
\cref{app:Hankel_Toeplitz_structures} contains visual representations of the structures explained in the main body of the paper.
Lastly, \cref{app:experiments} provides a full description of all the included experiments and the data used in them, with additional plots and results.

\section{Proof of~\cref{cor:polynomial} (use of~\cref{thm:2md} for Polynomial Basis Functions)}\label{app:polynomial}

Polynomial \glspl{bf}s (as used in for example~\citet{chen_parallelized_2018}) are defined along each dimension as
\begin{equation}
    \phi_{i_d}^{(d)}(x^{(d)})=(x^{(d)})^{i_d-1}
\end{equation}
By selecting $g_{i_d+j_d}(x^{(d)})=(x^{(d)})^{j_d+i_d-1}$, the product of two component \glspl{bf} becomes
\begin{equation}
    \phi_{i_d}^{(d)}(x^{(d)})\phi_{j_d}^{(d)}(x^{(d)})=(x^{(d)})^{i_d-1}(x^{(d)})^{j_d-1}=(x^{(d)})^{j_d+i_d-1}=g_{i_d+j_d},
\end{equation}
which shows that the conditions for \cref{thm:2md} is satisfied.

\section{Proof of~\cref{cor:complexexponential} (use of~\cref{thm:2md} for Complex Exponential Basis Functions)}\label{app:complexexponentialproof}

Complex exponential \glspl{bf} (as used in for example~\citet{novikov_exponential_2018}) are defined along each dimension as
\begin{equation}
    \phi_{i_d}^{(d)}(x^{(d)})=\exp(i\pi{j_d}x^{(d)})
\end{equation}
By selecting $g_{i_d+j_d}(x^{(d)})=\exp(i\pi(j_d+i_d)x^{(d)})$, the product of two component \glspl{bf} becomes
\begin{equation}
    \phi_{i_d}^{(d)}(x^{(d)})\phi_{j_d}^{(d)}(x^{(d)})=\exp(i\pi{i_d}x^{(d)})\exp(i\pi{j_d}x^{(d)})=\exp(i\pi({j_d+i_d})x^{(d)})=g_{i_d+j_d},
\end{equation}
which shows that the conditions for \cref{thm:2md} is satisfied.

\section{Proof of~\cref{cor:hgp} (use of~\cref{thm:3md} for Hilbert Space Basis Functions on a Rectangular Domain)}\label{app:hgprectangular}

Hilbert space \glspl{bf} on a rectangular domain are defined according to~\citep{solinHilbertSpaceMethods2020}
\begin{equation}
    \phi_i(x)=\prod_{d=1}^D \frac{1}{\sqrt{L_d}}\sin\left (\frac{\pi i_d (x_{d}+L_{d})}{2L_{d}}\right)=\prod_{d=1}^D\frac{1}{\sqrt{L_d}}\cos\left (\frac{\pi i_d (x_{d}+L_{d})}{2L_{d}}-\frac{\pi}{2}\right),
\end{equation}
where the indices $i={1,\hdots,M^D}$ has a one-to-one mapping with all possible combinations of the indices ${i_1, i_2, \hdots, i_D}$, given that each index $i_d=\{1,\hdots,D\}$.

This corresponds to defining the \glspl{bf} according to \cref{eq:kroenecker_BFs}, with each entry of $\mat{\phi}^{(d)}(x)$ defined as
\begin{equation*}
    {\phi}_{i_d}^{(d)}(x)=\tfrac{1}{\sqrt{L_{d}}}\sin\left (\frac{\pi i_d (x+L_{d})}{2L}\right )=\tfrac{1}{\sqrt{L_{d}}}\cos\left (\frac{\pi i_d (x+L_{d})}{2L_{d}}-\frac{\pi}{2}\right ),
\end{equation*}
where $i_d\in\{1,\hdots,m\}$. Define a linear function $\theta_{i_d}:\mathbb{R}\rightarrow\mathbb{R}$ as
\begin{equation*}
    \theta_{i_d}(x)=\frac{\pi {i_d} (x+L_{d})}{2L_{d}}-\frac{\pi}{2}.
\end{equation*}

Hence, the \glspl{bf} can be written as 
\begin{equation*}
    \mat{\phi}^{(d)}(x) = \frac{1}{\sqrt{L_d}}\begin{bmatrix}
        \cos(\theta_1)\\
        \vdots\\
        \cos(\theta_M)
    \end{bmatrix}.
\end{equation*}
The product $\mat{\phi}^{(d)}(x_n^{(d)})\mat{\phi}^{(d)}(x_n^{(d)})\trans$ is then given by
\begin{equation*}
    \mat{\phi}^{(d)}(x_n^{(d)})\mat{\phi}^{(d)}(x_n^{(d)})\trans = \frac{1}{L_d}\sum_{n=1}^N \begin{bmatrix}
        \cos(\theta_1)\cos(\theta_1) & \dots & \cos(\theta_1)\cos(\theta_M)\\
        \vdots & \ddots & \vdots\\
        \cos(\theta_M)\cos(\theta_M) & \dots & \cos(\theta_M)\cos(\theta_M)
    \end{bmatrix}.
\end{equation*}
Then, note that
\begin{equation}\label{eq:trig_identity}
\cos(u)\cos(v)=\frac{\cos(u+v)+\cos(u-v)}{2}.
\end{equation}
When we apply this to each entry in the matrix, we get that
\begin{equation*}
\begin{split}
    &\mat{\phi}^{(d)}(x_n^{(d)})\mat{\phi}^{(d)}(x_n^{(d)})\trans =\\
    &\frac{1}{2L_{d}}\Bigg(
    \underbrace{
    \sum_{n=1}^N 
    \begin{bmatrix}
        \cos(\theta_1+\theta_1) & \dots & \cos(\theta_1+\theta_M)\\
        \vdots & \ddots & \vdots \\
        \cos(\theta_M + \theta_1) & \dots & \cos(\theta_M + \theta_M)
    \end{bmatrix}}_{\triangleq \mat{G}^{d,(1)}}
    + \underbrace{\sum_{n=1}^N
    \begin{bmatrix}
        \cos(\theta_1-\theta_1) & \dots & \cos(\theta_1-\theta_M)\\
        \vdots & \ddots & \vdots \\
        \cos(\theta_M - \theta_1) & \dots & \cos(\theta_M - \theta_M)
    \end{bmatrix}}_{\triangleq \mat{G}^{d,(-1)}}
    \Bigg),
\end{split}
\end{equation*}
where $\mat{G}^{d,(1)}$ and $\mat{G}^{d,(-1)}$ are a Hankel and a Toeplitz matrix, respectively.
Then, since $\cos(\theta_{i_{d}} \pm \theta_{j_d}) = \pm\sin(\theta_{{i_d}\pm {j_d}})\triangleq g(i_d \pm j_d) = g(k_d)$, there exists a function $g(k_d)$ that fulfills conditions for \cref{thm:3md}.

\section{Proof of~\cref{cor:fourier} (use of~\cref{thm:3md} for Ordinary Fourier Features)}\label{app:ordinatyFourierFeatures}

Ordinary Fourier features for separable kernels are defined slightly differently from Hilbert Space \glspl{bf}~\citep{hensmanVariational2017}, in that they consider both a sine and a cosine function for each considered frequency. The set of \glspl{bf} are given by
\begin{equation}
\begin{split}
    \mat{\phi}^{(d)}(x)=&\begin{bmatrix}
\mat{\phi}^{(d)}_{\text{sin}}(x) & \mat{\phi}_{\text{cos}}^{(d)}(x)
\end{bmatrix}\\
=&\begin{bmatrix}
\sin(\Delta x) & \sin(2\Delta x) & \hdots & \sin(m_d\Delta) &
\cos(\Delta x) & \cos(2\Delta x) & \hdots & \cos(m_d\Delta)  
\end{bmatrix},
\end{split}
\end{equation}
where $\Delta$ determines the spacing of the Fourier features in the frequency domain. The precision matrix can therefore be expressed as
\begin{equation}
    \mat{\Phi}\trans\mat{\Phi}=\begin{bmatrix}
\mat{\Phi}_{\text{sin}}\trans\mat{\Phi}_{\text{sin}} & (\mat{\Phi}_{\text{cos}}\trans\mat{\Phi}_{\text{sin}})\trans\\ 
\mat{\Phi}_{\text{cos}}\trans\mat{\Phi}_{\text{sin}} & \mat{\Phi}_{\text{cos}}\trans\mat{\Phi}_{\text{cos}}
\end{bmatrix},
\end{equation}
and we can apply~\cref{thm:3md} directly to entries $\mat{\Phi}_{\sin}^\top\mat{\Phi}_{\sin}$ and $\mat{\Phi}_{\sin}^\top\mat{\Phi}_{\cos}$ to prove that each of these have a block Hankel--Toeplitz structure. 

For the matrix $\mat{\Phi}_{\sin}\trans\mat{\Phi}_{\sin}$, the product $\mat{\phi}_{\sin}(x_n^{(d)})\mat{\phi}_{\sin}(x_n^{(d)})\trans$ can be expanded as 
\begin{equation}
\begin{split}
    &\{ \mat{\phi}_{\sin}(x_n^{(d)})\mat{\phi}_{\sin}(x_n^{(d)})\trans \}_{i,j}=\sin(i\Delta x^{(d)})\sin(j\Delta x^{(d)})=\{\mat{G}^{(d),(-1)}\}_{i,j}+\{\mat{G}^{(d),(1)}\}_{i,j},
\end{split}
\end{equation}
where
\begin{equation}
    \{\mat{G}^{(d),(1)}\}_{i,j}=\cos(i\Delta x^{(d)}+j\Delta x^{(d)}) ,
\end{equation}
and
\begin{equation}
    \{\mat{G}^{(d),(-1)}\}_{i,j}=
\cos(\Delta x^{(d)}-\Delta x^{(d)}).
\end{equation}
The entry at row $i$ and column $j$ of $\mat{G}^{(d),(1)}$ can therefore be defined by the function $g(i+j)=-\cos(i+j)\Delta x^{(d)}$, and the entry at row $i$ and column $j$ of $\mat{G}^{(d),(-1)}$ is given by $-g(i-j)$, satisfying the requirements for~\cref{thm:3md}.

For the matrix $\mat{\Phi}_{\cos}\trans\mat{\Phi}_{\sin}$, the product $\mat{\phi}_{\cos}(x_n^{(d)})\mat{\phi}_{\sin}(x_n^{(d)})\trans$ can be expanded as 
\begin{equation}
\begin{split}
    \{\mat{\phi}_{\cos}(x_n^{(d)})\mat{\phi}_{\sin}(x_n^{(d)})\trans\}_{i,j}=
\cos(i \Delta x^{(d)})\sin(j \Delta x^{(d)}) 
=\{\mat{G}^{(d),(-1)}\}_{i,j}+\{\mat{G}^{(d),(1)}\}_{i,j};
\end{split}
\end{equation}
where
\begin{equation}
    \{\mat{G}^{(d),(1)}\}_{i,j}=\sin(i\Delta x^{(d)}+j\Delta x^{(d)}),
\end{equation}
and
\begin{equation}
    \{\mat{G}^{(d),(-1)}\}_{i,j}=
\sin(i\Delta x^{(d)}-j\Delta x^{(d)})
\end{equation}
The entry at row $i$ and column $j$ of $\mat{G}^{(d),(1)}$ can therefore be defined by the function $g(i+j)=-\cos((i+j)\Delta x^{(d)})$, and the entry at row $i$ and column $j$ of $\mat{G}^{(d),(-1)}$ is given by $-g(i-j)$, satisfying the requirements for~\cref{thm:3md}.

For the matrix $\mat{\Phi}_{\cos}\trans\mat{\Phi}_{\cos}$, the product $\mat{\phi}_{\cos}(x_n^{(d)})\mat{\phi}_{\cos}(x_n^{(d)})\trans$ can be expanded as 
\begin{equation}
\begin{split}
    &\{\mat{\phi}_{\cos}(x_n^{(d)})\mat{\phi}_{\cos}(x_n^{(d)})\trans\}_{i,j}=\cos(i\Delta x^{(d)})\cos(j\Delta x^{(d)})=
    \{\mat{G}^{(d),(-1)}\}_{i,j}+\{\mat{G}^{(d),(1)}\}_{i,j};
\end{split}
\end{equation}
where
\begin{equation}
    \{\mat{G}^{(d),(1)}\}_{i,j}=\cos(i\Delta x^{(d)}+j\Delta x^{(d)}),
\end{equation}
and
\begin{equation}
    \{\mat{G}^{(d),(-1)}\}_{i,j}=\cos(i\Delta x^{(d)}-j\Delta x^{(d)}) .
\end{equation}
The entry at row $i$ and column $j$ of $\mat{G}^{(d),(1)}$ can therefore be defined by the function $g(i+j)=\cos((i+j)\Delta x^{(d)})$, and the entry at row $i$ and column $j$ of $\mat{G}^{(d),(-1)}$ is given by $g(i-j)$. An important notion for this matrix which makes it different from $\mat{\Phi}_{\sin}\mat{\Phi}_{\sin}\trans$ and $\mat{\Phi}_{\cos}\mat{\Phi}_{\sin}\trans$ is that this does not exactly satisfy the criteria for~\cref{thm:3md}. The difference is that for the criteria to be exactly satisfied, entry $\{\mat{G}^{(d),(-1)}\}_{i,j}$ would have to be equal to $-g(i-j)$ rather than $g(i-j)$. However, by applying the proof of~\cref{thm:3md}, but now noticing that the entries of $\mat{C}=\mat{\Phi}_{\cos}\trans\mat{\Phi}_{\cos}$ can be expressed element-wise as
\begin{equation}
    \mat{C}_{i,j}=\sum_{p=1}^{2^D}\sum_{n=1}^N\prod_{d=1}^D g(i_d+e_p^{(d)}j_d),
\end{equation}
which allows us to use the tensor $\gamma$ as defined in~\cref{eq:gamma} to express each entry of $\mat{C}$ according to
\begin{equation}
\mat{C}_{i,j}=\sum_{p}^{2^D}\gamma_{i_1+e_p^{(1)}},\hdots,i_D+e_p^{(D)}.
\end{equation}

\section{Overview of Hankel, Toeplitz, and D-level Block Hankel--Toeplitz Matrix Structures}\label{app:Hankel_Toeplitz_structures}

An overview of Hankel, Toeplitz and D-level Block Hankel--Toeplitz matrix structures are given in~\cref{tab:Hankel_Toeplitz_structures}.

\begin{table}[]
\newcolumntype{M}[1]{>{\centering\arraybackslash}m{#1}}
\caption{An overview of the matrix structure and tensor representation for Hankel, Toeplitz, block Hankel, block Toeplitz and block Hankel matrices. The illustrations are examples of matrices with the property described in each row. The illustrations contain one square for each matrix entry, where the color of the square corresponds to the value. }
\label{tab:Hankel_Toeplitz_structures}
\vskip .15in
\centering
\begin{center}
\begin{small}
\begin{sc}
\begin{tabular}{m{.21\textwidth}cM{.125\textwidth}M{.15\textwidth}}
\toprule
Structure & Definition & Visual & Domain \\ \midrule
Hankel
& 
$\mat{H}=\begin{bmatrix}
\gamma_1 & \gamma_2 & \hdots & \gamma_m\\ 
\gamma_2 & \gamma_3 & \hdots & \gamma_{m+1}\\ 
\vdots & \vdots & \ddots & \vdots \\ 
\gamma_{m} & \gamma_{m+1} & \hdots & \gamma_{2m-1}
\end{bmatrix}$  
& 
\includegraphics[width=.125\textwidth]{GammaHankel1D.png}
& 
$\gamma\in\mathbb{R}^{K}$ 
\\ \midrule
Toeplitz 
&  
$\mat{T}=\begin{bmatrix}
\gamma_m & \hdots & \gamma_2 & \gamma_1\\ 
\gamma_{m+1} & \hdots & \gamma_3 & \gamma_2\\ 
\vdots & \iddots & \vdots & \vdots \\ 
\gamma_{2m-1} & \hdots & \gamma_{m+1} & \gamma_{m}
\end{bmatrix}$ 
& 
\includegraphics[width=0.125\textwidth]{GammaToeplitz1D.png}
& 
$\gamma\in\mathbb{R}^{K}$ 
\\\midrule
D-level block Hankel   
& 
$\mat{H}^{(D)}=\begin{bmatrix}
\mat{H}^{(D-1)}_{1} & \mat{H}_{2}^{(D-1)} & \hdots & \mat{H}_{m_D}^{(D-1)}\\ 
\mat{H}_{2}^{(D-1)} & \mat{H}_{3}^{(D-1)} & \hdots & \mat{H}_{m_D+1}^{(D-1)}\\ 
\vdots & \vdots & \ddots & \vdots \\ 
\mat{H}_{m_D}^{(D-1)} & \mat{H}_{m_D+1}^{(D-1)} & \hdots & \mat{H}_{2m_D-1}^{(D-1)}
\end{bmatrix}$
& 
\includegraphics[width=0.125\textwidth]{Gamma4.png}
&  
$\gamma\in\mathbb{R}^{K_1\times\hdots\times K_D}$ 
\\\midrule
D-level block Toeplitz & 
$\mat{T}^{(D)}=\begin{bmatrix}
\mat{T}_{m_D}^{(D-1)} & \hdots & \mat{T}_{2}^{(D-1)} & \mat{T}_{1}^{(D-1)}\\ 
\mat{T}_{m_D+1}^{(D-1)} & \hdots & \mat{T}_{2}^{(D-1)} & \mat{T}_{2}^{(D-1)}\\ 
\vdots & \iddots & \vdots & \vdots\\ 
\mat{T}^{(D-1)}_{2m_D-1} & \hdots  & \mat{T}_{m_D+1}^{(D-1)} & \mat{T}_{m_D}^{(D-1)}
\end{bmatrix}$ 
& 
\includegraphics[width=0.125\textwidth]{Gamma1.png}
& 
$\gamma\in\mathbb{R}^{K_1\times\hdots\times K_D}$ 
\\\midrule
D-level block \- Hankel--Toeplitz, level D is Hankel 
& 
$\mat{G}^{(D)}=\begin{bmatrix}
\mat{G}^{(D-1)}_{1} & \mat{G}_{2}^{(D-1)} & \hdots & \mat{G}_{m_D}^{(D-1)}\\ 
\mat{G}_{2}^{(D-1)} & \mat{G}_{3}^{(D-1)} & \hdots & \mat{G}_{m_D+1}^{(D-1)}\\ 
\vdots & \vdots & \ddots & \vdots \\ 
\mat{G}_{m_D}^{(D-1)} & \mat{G}_{m_D+1}^{(D-1)} & \hdots & \mat{G}_{2m_D-1}^{(D-1)}
\end{bmatrix}$ 
& 
\includegraphics[width=0.125\textwidth]{Gamma2.png}
& 
$\gamma\in\mathbb{R}^{K_1\times\hdots\times K_D}$ 
\\\midrule
D-level block \- Hankel--Toeplitz, level D is Toeplitz 
& 
$\mat{G}^{(D)}=\begin{bmatrix}
\mat{G}_{m_D}^{(D-1)} & \hdots & \mat{G}_{2}^{(D-1)} & \mat{G}_{1}^{(D-1)}\\ 
\mat{G}_{m_D+1}^{(D-1)} & \hdots & \mat{G}_{3}^{(D-1)} & \mat{G}_{2}^{(D-1)}\\ 
\vdots & \iddots & \vdots & \vdots\\ 
\mat{G}^{(D-1)}_{2m_D-1} & \hdots  & \mat{G}_{m_D+1}^{(D-1)} & \mat{G}_{m_D}^{(D-1)}
\end{bmatrix}$ 
& 
\includegraphics[width=0.125\textwidth]{Gamma3.png} 
& 
$\gamma\in\mathbb{R}^{K_1\times\hdots\times K_D}$
\\\bottomrule
\end{tabular}
\end{sc}
\end{small}
\end{center}
\end{table}

\section{Experiment Details}\label{app:experiments}
More specific details of the numerical experiments are provided here with additional visualizations and explanations.

\subsection{U.S. Precipitation Data}\label{app:precipitation}
\begin{figure}[t]
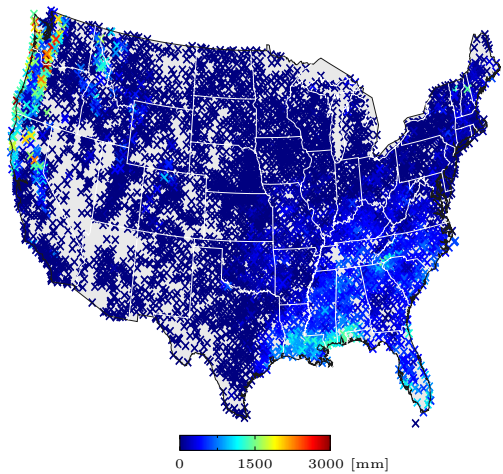

    \centering
    \includestandalone[]{\currfiledir /../Figures/precipitation/precipitation_data_orig}
    \caption{Raw data from precipitation data set. Each data point is visualized as a cross with color indicating the precipitation. The data set contains mostly low frequency content with some high frequency content apparent in the west coast as well as the south eastern parts.}
    \label{fig:precipitationdata}
\end{figure}

The precipitation data is two-dimensional with $N=5776$ data points first considered in \citep{vanhataloPrecipitation2008}.
We perform regression in the lat/lon domain and first center the data (but do not perform scaling) and use a simple squared-exponential kernel.
We optimized the \gls{mll} using GPJax \citep{PinderGPJAX2022} for both the \gls{hgp} as well as the standard \gls{gp} for $100$ iterations using Adam \citep{kingma2015Adam}.
The hyperparameters of the kernel and likelihood were initialized as $l=1,\sigma_{SE}^2=10$ and $\sigma_e=1$, where $l$ is the lengthscale, $\sigma_{SE}^2$ is the kernel variance and $\sigma_e$ the noise standard deviation.
Purely for visualization purposes, the inputs were projected to a local coordinate system given by CRS 5070 which are used for all of the plots.
The original data is plotted in \cref{fig:precipitationdata}.
The timing experiments were run using $m_d=5,10,\dots,65$ \glspl{bf} along each dimension, totaling between $M=25$ and $M=4225$ \glspl{bf}.

\subsection{Magnetic Field Mapping}\label{app:magneticfield}
\begin{figure}[t]
    \centering
    \begin{subfigure}{.48\columnwidth}
    \centering
        \begin{tikzpicture}

\begin{axis}[%
width=4.755cm,
height=5cm,
at={(0cm,0cm)},
scale only axis,
point meta min=-2,
point meta max=6,
axis on top,
xmin=7.438,
xmax=7.678,
xtick={7.45,  7.5, 7.55,  7.6, 7.65},
ymin=73.413,
ymax=73.479,
ytick={73.42, 73.44, 73.46},
xlabel={Latitude $({}^\circ)$},
ylabel={Longitude $({}^\circ)$},
label style={font=\scriptsize},
axis background/.style={fill=white},
xmajorgrids,
ymajorgrids,
grid style={opacity=0.5},
tick label style ={font=\tiny},
colormap={custom}{
  rgb255(0pt)=(0, 0, 131);
  rgb255(1pt)=(0, 0, 135);
  rgb255(2pt)=(0, 0, 139);
  rgb255(3pt)=(0, 0, 143);
  rgb255(4pt)=(0, 0, 147);
  rgb255(5pt)=(0, 0, 151);
  rgb255(6pt)=(0, 0, 155);
  rgb255(7pt)=(0, 0, 159);
  rgb255(8pt)=(0, 0, 163);
  rgb255(9pt)=(0, 0, 167);
  rgb255(10pt)=(0, 0, 171);
  rgb255(11pt)=(0, 0, 175);
  rgb255(12pt)=(0, 0, 179);
  rgb255(13pt)=(0, 0, 183);
  rgb255(14pt)=(0, 0, 187);
  rgb255(15pt)=(0, 0, 191);
  rgb255(16pt)=(0, 0, 195);
  rgb255(17pt)=(0, 0, 199);
  rgb255(18pt)=(0, 0, 203);
  rgb255(19pt)=(0, 0, 207);
  rgb255(20pt)=(0, 0, 211);
  rgb255(21pt)=(0, 0, 215);
  rgb255(22pt)=(0, 0, 219);
  rgb255(23pt)=(0, 0, 223);
  rgb255(24pt)=(0, 0, 227);
  rgb255(25pt)=(0, 0, 231);
  rgb255(26pt)=(0, 0, 235);
  rgb255(27pt)=(0, 0, 239);
  rgb255(28pt)=(0, 0, 243);
  rgb255(29pt)=(0, 0, 247);
  rgb255(30pt)=(0, 0, 251);
  rgb255(31pt)=(0, 0, 255);
  rgb255(32pt)=(0, 4, 255);
  rgb255(33pt)=(0, 8, 255);
  rgb255(34pt)=(0, 12, 255);
  rgb255(35pt)=(0, 16, 255);
  rgb255(36pt)=(0, 20, 255);
  rgb255(37pt)=(0, 24, 255);
  rgb255(38pt)=(0, 28, 255);
  rgb255(39pt)=(0, 32, 255);
  rgb255(40pt)=(0, 36, 255);
  rgb255(41pt)=(0, 40, 255);
  rgb255(42pt)=(0, 44, 255);
  rgb255(43pt)=(0, 48, 255);
  rgb255(44pt)=(0, 52, 255);
  rgb255(45pt)=(0, 56, 255);
  rgb255(46pt)=(0, 60, 255);
  rgb255(47pt)=(0, 64, 255);
  rgb255(48pt)=(0, 68, 255);
  rgb255(49pt)=(0, 72, 255);
  rgb255(50pt)=(0, 76, 255);
  rgb255(51pt)=(0, 80, 255);
  rgb255(52pt)=(0, 84, 255);
  rgb255(53pt)=(0, 88, 255);
  rgb255(54pt)=(0, 92, 255);
  rgb255(55pt)=(0, 96, 255);
  rgb255(56pt)=(0, 100, 255);
  rgb255(57pt)=(0, 104, 255);
  rgb255(58pt)=(0, 108, 255);
  rgb255(59pt)=(0, 112, 255);
  rgb255(60pt)=(0, 116, 255);
  rgb255(61pt)=(0, 120, 255);
  rgb255(62pt)=(0, 124, 255);
  rgb255(63pt)=(0, 128, 255);
  rgb255(64pt)=(0, 131, 255);
  rgb255(65pt)=(0, 135, 255);
  rgb255(66pt)=(0, 139, 255);
  rgb255(67pt)=(0, 143, 255);
  rgb255(68pt)=(0, 147, 255);
  rgb255(69pt)=(0, 151, 255);
  rgb255(70pt)=(0, 155, 255);
  rgb255(71pt)=(0, 159, 255);
  rgb255(72pt)=(0, 163, 255);
  rgb255(73pt)=(0, 167, 255);
  rgb255(74pt)=(0, 171, 255);
  rgb255(75pt)=(0, 175, 255);
  rgb255(76pt)=(0, 179, 255);
  rgb255(77pt)=(0, 183, 255);
  rgb255(78pt)=(0, 187, 255);
  rgb255(79pt)=(0, 191, 255);
  rgb255(80pt)=(0, 195, 255);
  rgb255(81pt)=(0, 199, 255);
  rgb255(82pt)=(0, 203, 255);
  rgb255(83pt)=(0, 207, 255);
  rgb255(84pt)=(0, 211, 255);
  rgb255(85pt)=(0, 215, 255);
  rgb255(86pt)=(0, 219, 255);
  rgb255(87pt)=(0, 223, 255);
  rgb255(88pt)=(0, 227, 255);
  rgb255(89pt)=(0, 231, 255);
  rgb255(90pt)=(0, 235, 255);
  rgb255(91pt)=(0, 239, 255);
  rgb255(92pt)=(0, 243, 255);
  rgb255(93pt)=(0, 247, 255);
  rgb255(94pt)=(0, 251, 255);
  rgb255(95pt)=(0, 255, 255);
  rgb255(96pt)=(4, 255, 251);
  rgb255(97pt)=(8, 255, 247);
  rgb255(98pt)=(12, 255, 243);
  rgb255(99pt)=(16, 255, 239);
  rgb255(100pt)=(20, 255, 235);
  rgb255(101pt)=(24, 255, 231);
  rgb255(102pt)=(28, 255, 227);
  rgb255(103pt)=(32, 255, 223);
  rgb255(104pt)=(36, 255, 219);
  rgb255(105pt)=(40, 255, 215);
  rgb255(106pt)=(44, 255, 211);
  rgb255(107pt)=(48, 255, 207);
  rgb255(108pt)=(52, 255, 203);
  rgb255(109pt)=(56, 255, 199);
  rgb255(110pt)=(60, 255, 195);
  rgb255(111pt)=(64, 255, 191);
  rgb255(112pt)=(68, 255, 187);
  rgb255(113pt)=(72, 255, 183);
  rgb255(114pt)=(76, 255, 179);
  rgb255(115pt)=(80, 255, 175);
  rgb255(116pt)=(84, 255, 171);
  rgb255(117pt)=(88, 255, 167);
  rgb255(118pt)=(92, 255, 163);
  rgb255(119pt)=(96, 255, 159);
  rgb255(120pt)=(100, 255, 155);
  rgb255(121pt)=(104, 255, 151);
  rgb255(122pt)=(108, 255, 147);
  rgb255(123pt)=(112, 255, 143);
  rgb255(124pt)=(116, 255, 139);
  rgb255(125pt)=(120, 255, 135);
  rgb255(126pt)=(124, 255, 131);
  rgb255(127pt)=(128, 255, 128);
  rgb255(128pt)=(131, 255, 124);
  rgb255(129pt)=(135, 255, 120);
  rgb255(130pt)=(139, 255, 116);
  rgb255(131pt)=(143, 255, 112);
  rgb255(132pt)=(147, 255, 108);
  rgb255(133pt)=(151, 255, 104);
  rgb255(134pt)=(155, 255, 100);
  rgb255(135pt)=(159, 255, 96);
  rgb255(136pt)=(163, 255, 92);
  rgb255(137pt)=(167, 255, 88);
  rgb255(138pt)=(171, 255, 84);
  rgb255(139pt)=(175, 255, 80);
  rgb255(140pt)=(179, 255, 76);
  rgb255(141pt)=(183, 255, 72);
  rgb255(142pt)=(187, 255, 68);
  rgb255(143pt)=(191, 255, 64);
  rgb255(144pt)=(195, 255, 60);
  rgb255(145pt)=(199, 255, 56);
  rgb255(146pt)=(203, 255, 52);
  rgb255(147pt)=(207, 255, 48);
  rgb255(148pt)=(211, 255, 44);
  rgb255(149pt)=(215, 255, 40);
  rgb255(150pt)=(219, 255, 36);
  rgb255(151pt)=(223, 255, 32);
  rgb255(152pt)=(227, 255, 28);
  rgb255(153pt)=(231, 255, 24);
  rgb255(154pt)=(235, 255, 20);
  rgb255(155pt)=(239, 255, 16);
  rgb255(156pt)=(243, 255, 12);
  rgb255(157pt)=(247, 255, 8);
  rgb255(158pt)=(251, 255, 4);
  rgb255(159pt)=(255, 255, 0);
  rgb255(160pt)=(255, 251, 0);
  rgb255(161pt)=(255, 247, 0);
  rgb255(162pt)=(255, 243, 0);
  rgb255(163pt)=(255, 239, 0);
  rgb255(164pt)=(255, 235, 0);
  rgb255(165pt)=(255, 231, 0);
  rgb255(166pt)=(255, 227, 0);
  rgb255(167pt)=(255, 223, 0);
  rgb255(168pt)=(255, 219, 0);
  rgb255(169pt)=(255, 215, 0);
  rgb255(170pt)=(255, 211, 0);
  rgb255(171pt)=(255, 207, 0);
  rgb255(172pt)=(255, 203, 0);
  rgb255(173pt)=(255, 199, 0);
  rgb255(174pt)=(255, 195, 0);
  rgb255(175pt)=(255, 191, 0);
  rgb255(176pt)=(255, 187, 0);
  rgb255(177pt)=(255, 183, 0);
  rgb255(178pt)=(255, 179, 0);
  rgb255(179pt)=(255, 175, 0);
  rgb255(180pt)=(255, 171, 0);
  rgb255(181pt)=(255, 167, 0);
  rgb255(182pt)=(255, 163, 0);
  rgb255(183pt)=(255, 159, 0);
  rgb255(184pt)=(255, 155, 0);
  rgb255(185pt)=(255, 151, 0);
  rgb255(186pt)=(255, 147, 0);
  rgb255(187pt)=(255, 143, 0);
  rgb255(188pt)=(255, 139, 0);
  rgb255(189pt)=(255, 135, 0);
  rgb255(190pt)=(255, 131, 0);
  rgb255(191pt)=(255, 128, 0);
  rgb255(192pt)=(255, 124, 0);
  rgb255(193pt)=(255, 120, 0);
  rgb255(194pt)=(255, 116, 0);
  rgb255(195pt)=(255, 112, 0);
  rgb255(196pt)=(255, 108, 0);
  rgb255(197pt)=(255, 104, 0);
  rgb255(198pt)=(255, 100, 0);
  rgb255(199pt)=(255, 96, 0);
  rgb255(200pt)=(255, 92, 0);
  rgb255(201pt)=(255, 88, 0);
  rgb255(202pt)=(255, 84, 0);
  rgb255(203pt)=(255, 80, 0);
  rgb255(204pt)=(255, 76, 0);
  rgb255(205pt)=(255, 72, 0);
  rgb255(206pt)=(255, 68, 0);
  rgb255(207pt)=(255, 64, 0);
  rgb255(208pt)=(255, 60, 0);
  rgb255(209pt)=(255, 56, 0);
  rgb255(210pt)=(255, 52, 0);
  rgb255(211pt)=(255, 48, 0);
  rgb255(212pt)=(255, 44, 0);
  rgb255(213pt)=(255, 40, 0);
  rgb255(214pt)=(255, 36, 0);
  rgb255(215pt)=(255, 32, 0);
  rgb255(216pt)=(255, 28, 0);
  rgb255(217pt)=(255, 24, 0);
  rgb255(218pt)=(255, 20, 0);
  rgb255(219pt)=(255, 16, 0);
  rgb255(220pt)=(255, 12, 0);
  rgb255(221pt)=(255, 8, 0);
  rgb255(222pt)=(255, 4, 0);
  rgb255(223pt)=(255, 0, 0);
  rgb255(224pt)=(251, 0, 0);
  rgb255(225pt)=(247, 0, 0);
  rgb255(226pt)=(243, 0, 0);
  rgb255(227pt)=(239, 0, 0);
  rgb255(228pt)=(235, 0, 0);
  rgb255(229pt)=(231, 0, 0);
  rgb255(230pt)=(227, 0, 0);
  rgb255(231pt)=(223, 0, 0);
  rgb255(232pt)=(219, 0, 0);
  rgb255(233pt)=(215, 0, 0);
  rgb255(234pt)=(211, 0, 0);
  rgb255(235pt)=(207, 0, 0);
  rgb255(236pt)=(203, 0, 0);
  rgb255(237pt)=(199, 0, 0);
  rgb255(238pt)=(195, 0, 0);
  rgb255(239pt)=(191, 0, 0);
  rgb255(240pt)=(187, 0, 0);
  rgb255(241pt)=(183, 0, 0);
  rgb255(242pt)=(179, 0, 0);
  rgb255(243pt)=(175, 0, 0);
  rgb255(244pt)=(171, 0, 0);
  rgb255(245pt)=(167, 0, 0);
  rgb255(246pt)=(163, 0, 0);
  rgb255(247pt)=(159, 0, 0);
  rgb255(248pt)=(155, 0, 0);
  rgb255(249pt)=(151, 0, 0);
  rgb255(250pt)=(147, 0, 0);
  rgb255(251pt)=(143, 0, 0);
  rgb255(252pt)=(139, 0, 0);
  rgb255(253pt)=(135, 0, 0);
  rgb255(254pt)=(131, 0, 0);
  rgb255(255pt)=(128, 0, 0);
}
,
colorbar horizontal,
colorbar style={height=0.2cm, 
width=4.75cm, 
major tick length=0.15em, 
xticklabel pos=top,
xtick={0, 2, 4, 6},
extra x ticks={-2},
extra x tick style={font=\tiny, xticklabel style={yshift=-.025cm}},
every x tick/.style={black},
tick label style={font=\tiny},
minor tick length=0.1em,
minor tick num=1,
at={(parent axis.north)},
anchor=south,
yshift=.5cm},
]
\addplot [forget plot] graphics [xmin=7.438, xmax=7.678, ymin=73.413, ymax=73.479] {\currfiledir magmeas-1.png};
\end{axis}
\end{tikzpicture}%
        \caption{Raw magnetic field measurements for the underwater magnetic field data. The plotted data is subsampled to 100\textsuperscript{th} of the data for visualization purposes.}
        \label{fig:magneticfieldmeas}
    \end{subfigure}\hfill
    \begin{subfigure}{.48\columnwidth}
    \centering
        \begin{tikzpicture}

\begin{axis}[%
width=4.755cm,
height=5cm,
at={(0cm,0cm)},
scale only axis,
axis on top,
xmin=7.438,
xmax=7.678,
xtick={7.45,  7.5, 7.55,  7.6, 7.65},
ymin=73.4149,
ymax=73.479,
ytick={73.42, 73.44, 73.46},
axis background/.style={fill=white},
xmajorgrids,
ymajorgrids,
grid style={opacity=0.5},
tick label style ={font=\tiny},
legend columns=2,
legend style={
at={(0.5, 1.02)},
anchor=south,
draw=none,
fill=black!02,
font=\scriptsize,
column sep=.5em,
},
xlabel={Latitude $({}^\circ)$},
ylabel={Longitude $({}^\circ)$},
label style={font=\scriptsize},
]
\addplot [forget plot] graphics [xmin=7.438, xmax=7.678, ymin=73.4149, ymax=73.479] {\currfiledir magdatadivision-1.png};
\addlegendimage{only marks, mark=square*, color={rgb:red,85;green,168;blue,104}, line width=1pt}
\addlegendentry[color=black]{Training data};
\addlegendimage{only marks, mark=square*, color={rgb:red,196;green,78;blue,82}, line width=1pt}
\addlegendentry[color=black]{Test data};
\end{axis}
\end{tikzpicture}%
        \caption{Data divided into training and test set. Deterministically split to ensure ensure that the lengthscale is captured properly in the training data.}
        \label{fig:magneticfielddatasplit}
    \end{subfigure}
    \caption{Magnetic field training data and test data. The data is roughly 50/50 split between training and test set. Both training and test data are normalized only by the mean and standard deviation of the training data.}
    \label{fig:magfieldappendix}
\end{figure}
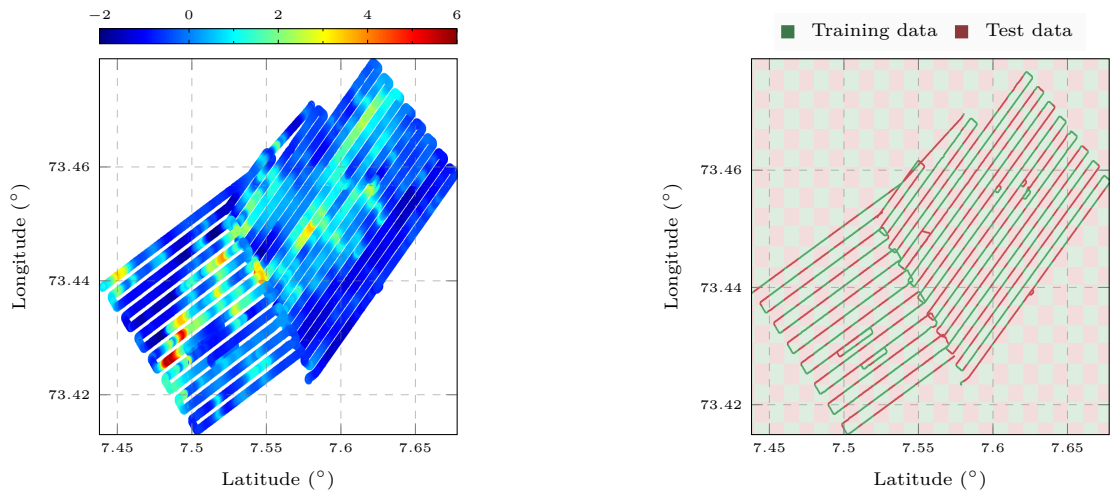

The magnetic field data has $N\approx 1.4~\text{million}$ data points and is collected with an underwater vehicle outside the Norwegian coast. The data used were collected by MARMINE/NTNU research cruise funded by the Research Council of Norway (Norges Forskningsråd, NFR) Project No. 247626/O30 and associated industrial partners. Ocean Floor Geophysics provided magnetometer that was used for magnetic data acquisition and pre‐processed the magnetic data. The data was later split into a training set and test set, at a roughly $50\%$ split. 
The nature of the data split is visualized in \cref{fig:magneticfielddatasplit}.
However, in practice, we selected the width of the test squares and the training squares smaller than the one displayed in the illustration and they are merely that big for visualization purposes.
The illustration displays squares that are $0.01$ latitudinal degrees wide and $0.03$ longitudinal degrees tall, corresponding to approximately $\SI{1}{\kilo\meter}$ in Cartesian coordinates in this area. 
The split we actually used was squares which were $0.001$ latitudinal degrees wide and $0.003$ longitudinal degrees tall, corresponding to approximately $\SI{100}{\meter}$ in both directions in Cartesian coordinates in that area. 
\gls{gp} regression with a squared exponential kernel is able to extrapolate for approximately one lengthscale, but will not necessarily give a very informative prediction one or two lengthscales away from the nearest measurement. 
Although we do not know the lengthscale that would optimize the likelihood of the data before using training data to find it, we see from a zoomed-in version of \cref{fig:magneticfielddata} approximately how fast the magnetic field is changing across the spatial dimension and use this to make a reasonable guess at the distance we expect a well-tuned \gls{gp} to be able to extrapolate the learned magnetic field. 
We then project the data into a local coordinate system using WGS84 and perform regression in Cartesian coordinates.
We center and standardize the data with the mean and standard deviation of the training data.
A squared-exponential kernel was used and we optimize the \gls{mll} in GPJax \citep{PinderGPJAX2022} using Adam \citep{kingma2015Adam} for $100$ iterations to find hyperparameters.
The hyperparameters were initialized as $l=\SI{200}{\meter},\sigma_{\text{SE}}^2=1$ and $\sigma_\text{y}^2=1$.
The resulting hyperparameters were $l_{SE}=190$, $\sigma_{\text{y}}^2=0.0675$, $\sigma_{\text{SE}}^2=7.15$.

\end{document}